\documentclass[10pt,twocolumn,letterpaper]{article}

\usepackage{iccv}
\usepackage{times}
\usepackage{epsfig}
\usepackage{graphicx}
\usepackage{amsmath}
\usepackage{amssymb}

\usepackage[accsupp]{axessibility} 

\usepackage[pagebackref=true,breaklinks=true,letterpaper=true,colorlinks,bookmarks=false]{hyperref}
\usepackage{subcaption}
\usepackage{graphicx}
\usepackage{amsmath}
\usepackage{amssymb}
\usepackage{booktabs}
\usepackage{comment}
\usepackage{multirow}
\newcommand{\argmin}{\arg\!\min} 
\usepackage[capitalize]{cleveref}
\crefname{section}{Sec.}{Secs.}
\Crefname{section}{Section}{Sections}
\Crefname{table}{Table}{Tables}
\crefname{table}{Tab.}{Tabs.}

\iccvfinalcopy 

\def\iccvPaperID{12055} 

\ificcvfinal\fi

\begin{document}

\title{EGformer: Equirectangular Geometry-biased Transformer for 360 Depth Estimation}

\author{Ilwi Yun\textsuperscript{\rm 1}
\and
Chanyong Shin\textsuperscript{\rm 2}
\and
Hyunku Lee\textsuperscript{\rm 1}
\and
Hyuk-Jae Lee\textsuperscript{\rm 1}
\and 
Chae Eun Rhee\textsuperscript{\rm 2} 
\and
\textsuperscript{\rm 1} Seoul National University, Seoul, Republic of Korea   
\and
 \textsuperscript{\rm 2}Inha University, Incheon, Republic of Korea 
\and 
{\tt\small yuniw@capp.snu.ac.kr}, {\tt\small scyongg@inha.edu}, {\tt\small hyunku.lee@snu.ac.kr}, \\ {\tt\small hjlee@capp.snu.ac.kr}, {\tt\small chae.rhee@inha.ac.kr} 
}

\maketitle
\ificcvfinal\thispagestyle{empty}\fi

\begin{abstract}
Estimating the depths of equirectangular (\ie, 360$^\circ$) images (EIs) is challenging given the distorted 180$^\circ$ $\times$ 360$^\circ$ field-of-view, which is hard to be addressed via convolutional neural network (CNN). Although a transformer with global attention achieves significant improvements over CNN for EI depth estimation task, it is computationally inefficient, which raises the need for transformer with local attention. However, to apply local attention successfully for EIs, a specific strategy, which addresses distorted equirectangular geometry and limited receptive field simultaneously, is required. Prior works have only cared either of them, resulting in unsatisfactory depths occasionally. 
 In this paper, we propose an equirectangular geometry-biased transformer termed EGformer. While limiting the computational cost and the number of network parameters, EGformer enables the extraction of the equirectangular geometry-aware local attention with a large receptive field. To achieve this, we actively utilize the equirectangular geometry as the bias for the local attention instead of struggling to reduce the distortion of EIs.
As compared to the most recent EI depth estimation studies, 
the proposed approach yields the best depth outcomes overall with the lowest computational cost and the fewest parameters, demonstrating the effectiveness of the proposed methods.
\end{abstract}

\section{Introduction}

\label{sec:introduction}
Estimating the depths of equirectangular (\ie, 360$^\circ$) images (EIs) can be challenging because such images have a 180$^\circ$ $\times$ 360$^\circ$ wide field-of-view (FoV) with distortion. Images with distorted wide FoV often requires a global view for proper image processing \cite{geo_depth,Slicenet,joint_360depth}. Such circumstances strongly require a large receptive field for accurate depth estimations of EIs, which is hard to be achieved via convolutional neural network (CNN).  

Considering the receptive field, the vision transformer (ViT) \cite{vision_transformer} may be the best option for equirectangular depth estimations. ViT has advantages over a CNN in that the attention is extracted in a global manner. A wide FoV can be addressed through global attention, and the effectiveness of this approach has been demonstrated \cite{joint_360depth}. However, in terms of the computational cost, this global mechanism makes ViT inappropriate for application to EIs. The computational cost of global attention is quadratic with respect to the input resolutions \cite{SwinT} which raises the need for local attention \cite{SwinT,Swin_v2,Cswin,dilated_attention,dilated_attention2}.

\begin{figure}[t!]
\centering
\begin{minipage}{.245\textwidth}
\begin{subfigure}{\linewidth}
\includegraphics[width=.98\linewidth]{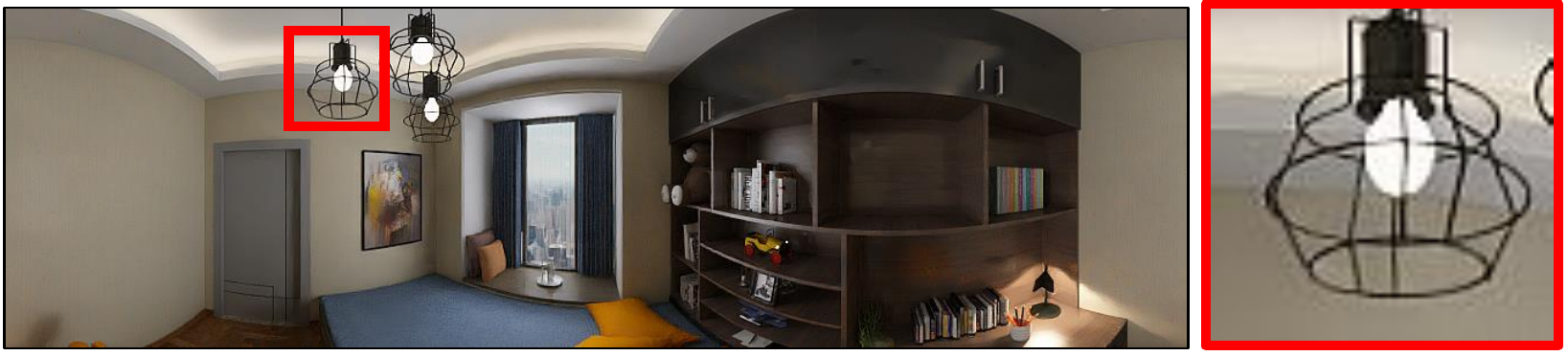}
\end{subfigure}
\subcaption{Input \cite{structure3d}}
\end{minipage}%
\begin{minipage}{.245\textwidth}
\begin{subfigure}{\linewidth}
\includegraphics[width=.98\linewidth]{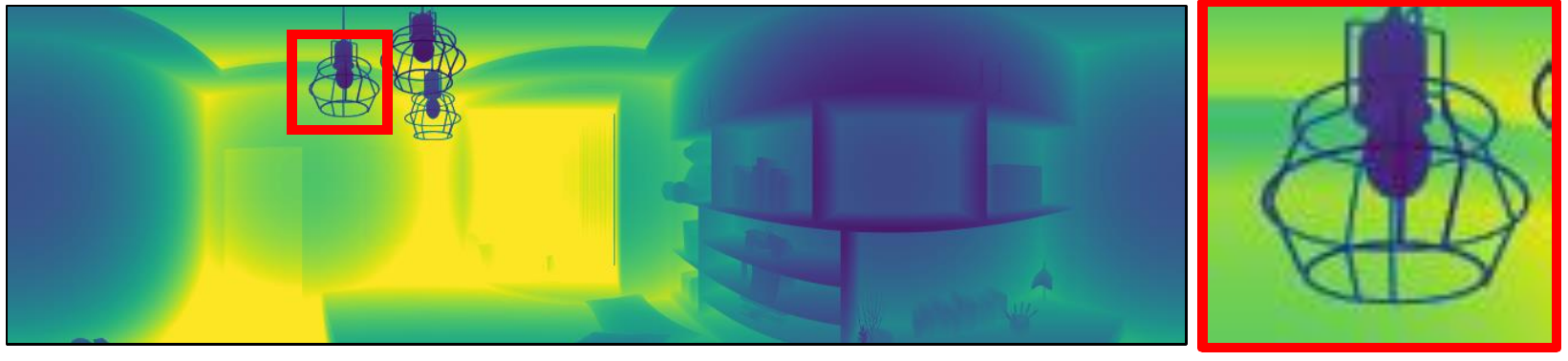}
\end{subfigure}
\subcaption{Ground truth \cite{structure3d}}
\end{minipage}%

\begin{minipage}{.245\textwidth}
\captionsetup{justification=centering}
\begin{subfigure}{\linewidth}
\includegraphics[width=.98\linewidth]{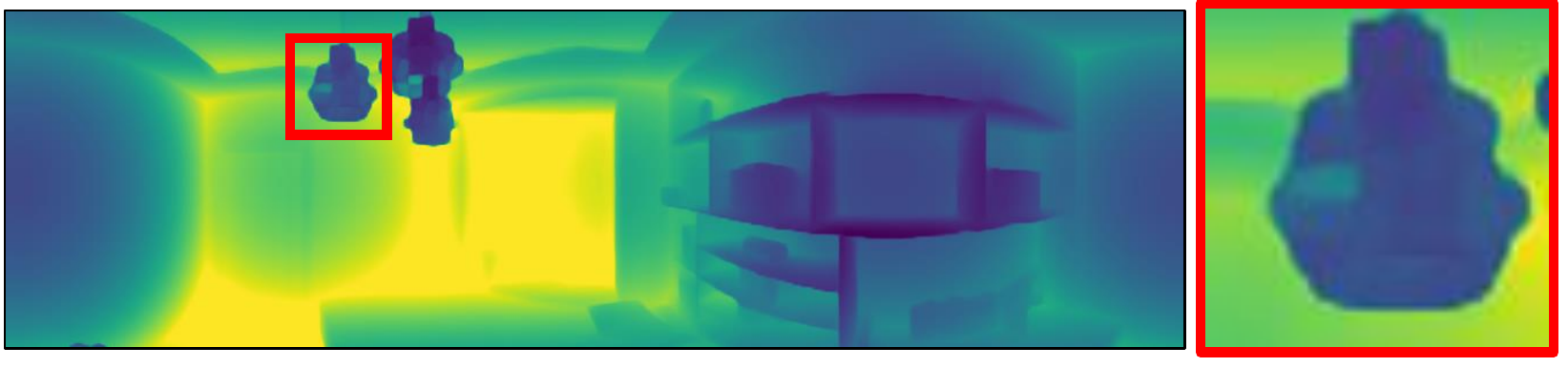}
\end{subfigure}
\subcaption{Panoformer \cite{panoformer}\\ (77.7GFlops, 20.4M params)}
\end{minipage}%
\begin{minipage}{.245\textwidth}
\captionsetup{justification=centering}
\begin{subfigure}{\linewidth}
\includegraphics[width=.98\linewidth]{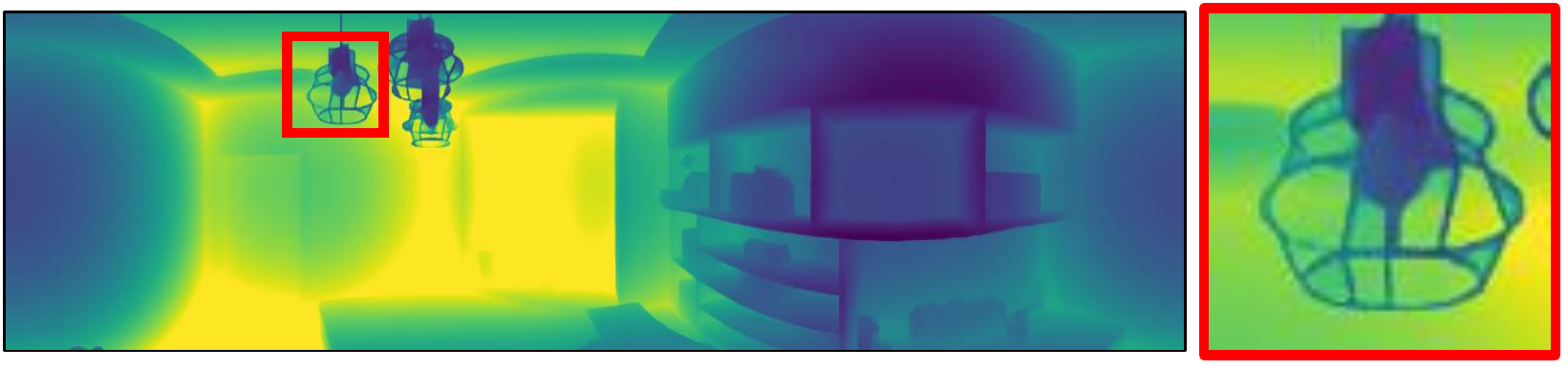}
\end{subfigure}
\subcaption{EGformer (73.9GFlops,15.4M params)}
\end{minipage}%

\caption{By utilizing the equirectangular geometry as the bias, EGformer efficiently enables the extraction of the equirectangular geometry-aware local attention with a large receptive field, yielding much more accurate depths with the lowest computational cost and the fewest parameters as compared to the results in the most recent studies.}
\label{fig:Intro}
\end{figure}

Unfortunately, applying local attention to EIs is a non-trivial problem because the distorted geometry and local receptive field should be addressed at the same time. Due to the non-uniform geometry of EIs, each local attention should be extracted differently while considering the equirectangular geometry. Therefore, local attention for general vision tasks cannot yield satisfactory performance outcomes for EIs, which demands specific strategies. However, even if the distortion of EIs can be addressed via proper strategies as proposed recently by Panoformer \cite{panoformer}, there still remains a fundamental limitation of local attention: the limited receptive field. 
To enlarge the receptive field, various hand-crafted or data-adaptive sparse patterns have been proposed for local windows with a hierarchical type of network architecture \cite{pyramidformer,pyramidformer2,SwinT,Cswin,DAT,panoformer}. However, such repeated local operations cannot fundamentally substitute for a global operator \cite{non_local}, increasing both the computational cost and the number of network parameters required for a plausible quality of the depth.

As described above, dealing efficiently with equirectangular geometry and a limited receptive field via local attention appears to be challenging, yet one important fact has been overlooked: \textbf{the equirectangular geometry is known beforehand}. For various vision tasks, it has been shown that a structural prior can boost the performances efficiently \cite{geo_depth,position_gan,HA_net,structural_sonar,landmark_face,struct_inpainting}. For example, based on the prior knowledge that cars cannot fly up in the sky in urban scenes, HA-Net \cite{HA_net} improves segmentation performance outcomes at a negligible computational overhead by imposing different importance levels on the encoded features according to their vertical positions. Inspired by those studies, we come up with the idea of offsetting the limitations of local attention via a structural prior of EIs.

In this paper, we propose an equirectangular geometry-biased transformer, termed EGformer, which actively utilizes the equirectangular geometry as the bias for inter- and intra-local windows. Through this, while limiting the computational cost and the number of network parameters, EGformer enables the extraction of the equirectangular geometry-aware local attention with a large receptive field. EGformer consists of three main proposals: equirectangular relative position embedding (ERPE), distance-based attention score (DAS) and equirectangular-aware attention rearrangement (EaAR). 
Specifically, ERPE and DAS impose geometry bias onto the elements within the local window, allowing for consideration of the equirectangular geometry when extracting the local attention. Meanwhile, EaAR imposes the geometry bias on the local window. This enables each local window to interact with other local windows indirectly, thereby enlarging the receptive field. Compared to the most recent studies of EI depth estimations, EGformer yields the best depth outcomes overall with the lowest computational cost and the fewest parameters, demonstrating the effectiveness of the proposed method.

\section{Background and related work}
\label{sec:related_work}

\subsection{Equirectangular geometry}
\label{Sec:equi_geometry}
\begin{figure}[h!]
\centering
\begin{minipage}{.49\textwidth}
\begin{subfigure}{\linewidth}
\includegraphics[width=.98\linewidth]{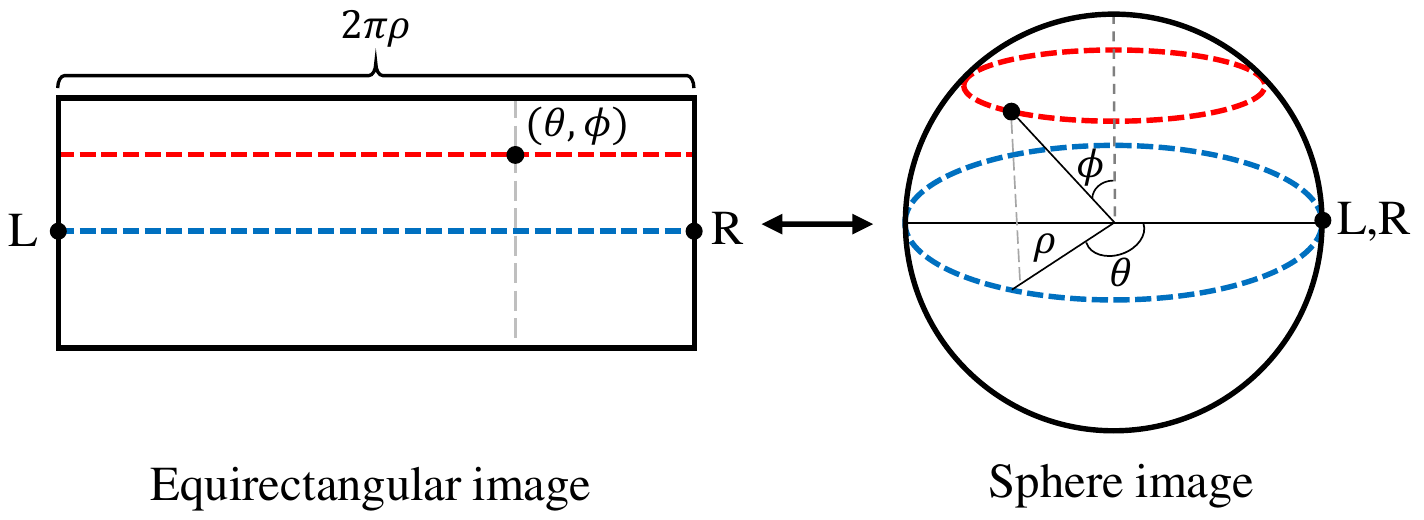}
\end{subfigure}
\end{minipage}%
\caption{Conversion between EI and sphere image}
\label{fig:equirectangular_geometry}
\end{figure}

As shown in Figure \ref{fig:equirectangular_geometry}, EIs are constructed by projecting a sphere image onto two-dimensional (2D) plane, and vice versa. Therefore, spherical coordinates are used for EIs, and each pixel location is represented through $(\rho, \theta ,\phi)$, where $\theta \in (0,2\pi)$, $\phi \in (0,\pi)$. Spherical coordinates can be converted to Cartesian coordinates ($X,Y,Z$) via Eq.(\ref{Eq:conversion}).

\begin{equation}
\label{Eq:conversion}
\begin{cases}
X = \rho \cdot \sin(\phi) \cdot \cos(\theta) \\
Y = \rho \cdot \sin(\phi) \cdot \sin(\theta) \\
Z = \rho \cdot \cos(\phi)
\end{cases}
\end{equation}

\paragraph{Structural prior}
Due to the equirectangular geometry, EIs have some distinct characteristics which should be further considered for proper image processing. Below are several examples.
\textbf{First}, the information density of EIs differ according to the locations.
 The information density is low around near $\phi=0,\pi$, while it is high at $\phi={\pi \over 2}$. The red and blue lines in Figure \ref{fig:equirectangular_geometry} represent the differences in the information density. Despite having the same number of pixels, the red and blue lines in EIs contain different amount of information, as shown in the sphere image of Figure \ref{fig:equirectangular_geometry}. Therefore, even with equal local window sizes, there exist differences in information quantity according to the locations of local window. 
\textbf{Second}, EIs are cyclic. In other words, the left and right ends of EIs are actually connected although they appear to be separated in EIs. The L and R points in Figure \ref{fig:equirectangular_geometry} visualize this characteristic. As the worst case, a single object is often split into left and right ends of EIs, requiring specific strategy \cite{EBS}.

\begin{figure*}[t!]
\centering
\begin{minipage}{.73\textwidth}
\begin{subfigure}{\linewidth}
\includegraphics[width=.98\linewidth]{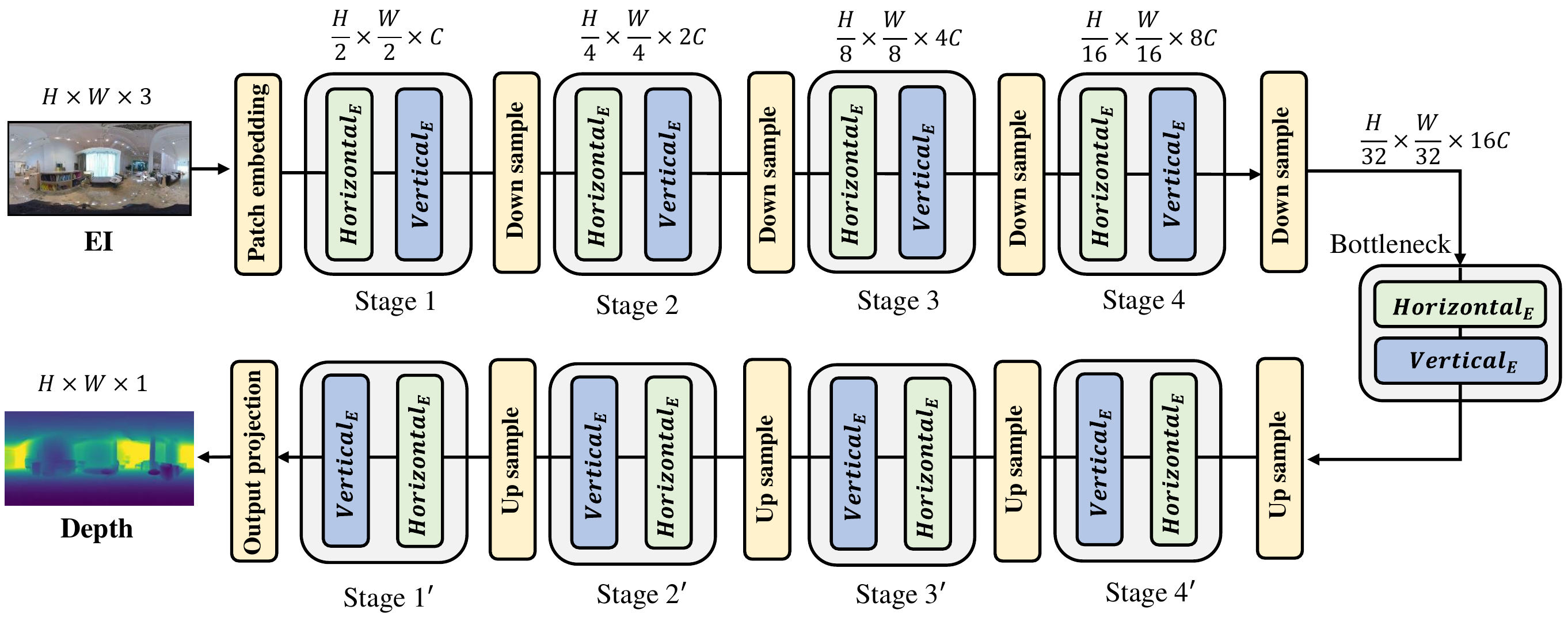}
\end{subfigure}
\end{minipage}%
\begin{minipage}{.30\textwidth}
\begin{subfigure}{\linewidth}
\includegraphics[width=.98\linewidth]{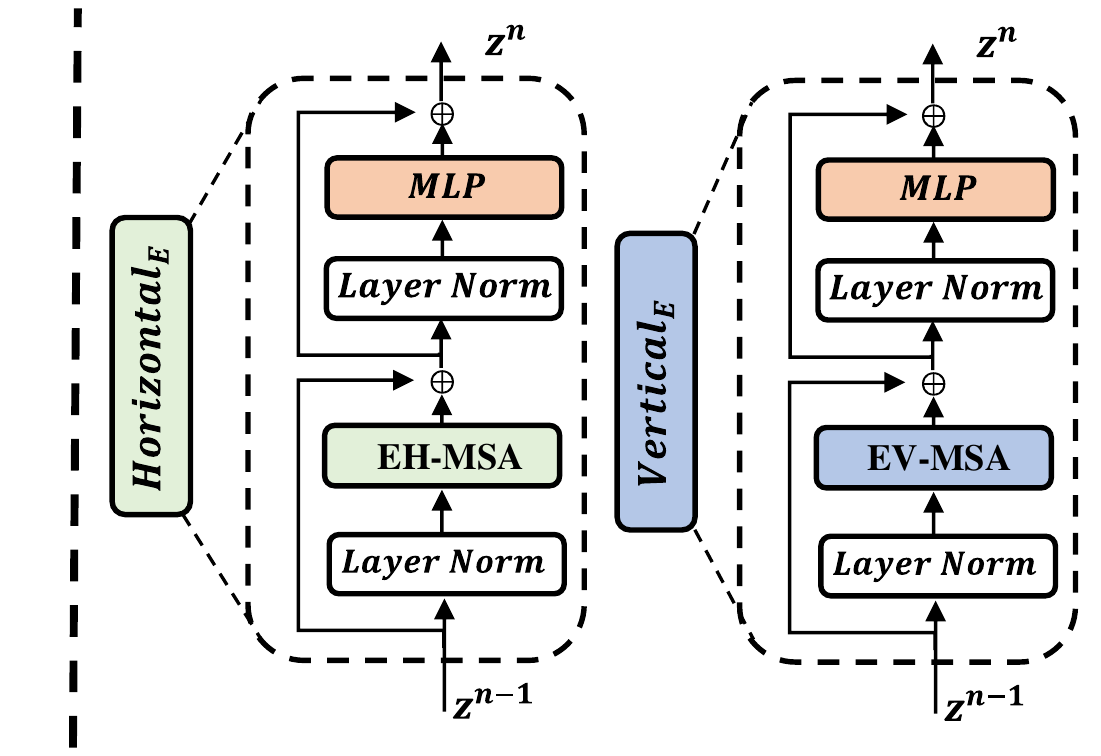}
\end{subfigure}
\end{minipage}%

\caption{The network architecture of EGformer variants (left) and EGformer transformer block (right). Stage n and Stage n$'$ indicates encoder and decoder, respectively. For better visualization, skip connection\cite{Unet} between encoder and decoder is omitted here, and more details are included in \textbf{Technical Appendix}. $Horizontal_E$ and $Vertical_E$ represents the proposed transformer block of EGformer (Section \ref{Sec:attention_block}), which comprise equirectangular-aware vertical and horizontal multi-head self-attention (\ie, EH-MSA,EV-MSA) (Section \ref{Sec:E-MSA}). }
\label{fig:overall_architecture}
\end{figure*} 

\subsection{Transformer for vision tasks}
 Compared to a CNN, ViT \cite{vision_transformer} possesses a global receptive field, which is highly beneficial for various vision tasks \cite{crossvit,TNT,T2T_ViT,trans_depth,DPT,depthformer}.
However, due to the high computational cost of global attention, several studies have focused on utilizing the local attention based on hierarchical architecture \cite{pyramidformer,pyramidformer2}.
 The Swin Transformer (SwinT) \cite{SwinT} proposes square-shaped local attention and the associated shifting mechanism, and Deformable attention transformer \cite{DAT} further improves SwinT through deformable attention inspired by the deformable convolution \cite{deformable_conv,deformable_conv2}. 
 However, square-shaped local attention with a hierarchical architecture enlarges the receptive field too slowly. To alleviate this, the CSwin transformer (CSwinT) \cite{Cswin} proposes the use of horizontal and vertical local attention in parallel, and Dilateformer \cite{dilated_attention2} proposes dilated attention inspired by dilated convolution \cite{dilated_convolution}. Nevertheless, both remain limited in that the receptive field is bounded to the sizes of the local windows.

\subsection{360 monocular depth estimation}
\label{Sec:360_detph}
The topics of equirectangular depth estimation studies mostly fall into two categories: Dealing with equirectangular geometry or dealing with insufficeint data. To address the distortion of EIs, several studies \cite{bifuse,bifuse_plus} utilize cubemap projections with certain padding schemes \cite{cubepadding}. Convolution kernels considering equirectangular geometry \cite{spherical_cnns,spherical_convolution} have also been studied.  Instead of addressing the distortion directly, some studies utilize the equirectangular geometry to improve the performance. Based on the finding that the geometric structures of EIs are embedded along the vertical direction \cite{360align}, Hohonet \cite{hohonet} and SliceNet \cite{Slicenet} propose to process EIs in a vertical direction. 
 Jin \etal \cite{geo_depth} and Zeng \etal \cite{jointdepth} show that some prior knowledge of geometric structure of EIs can boost the performances further.
  Meanwhile, due to the distorted and wide FoV, acquiring ground truth equirectangular depths is extremely difficult, resulting in lack of data \cite{Lowcost,omnidepth,joint_360depth}. Therefore, some studies have attempted to address data insufficiency through self-supervised learning \cite{360self,svsyn,EBS,360sdnet} or transfer learning \cite{joint_360depth}. 
  Recently, inspired by the success of ViT \cite{vision_transformer}, there have been several attempts to apply a transformer to equirectangular depth estimations. Yun \etal \cite{joint_360depth} demonstrated that global attention can effectively handle the wide FoV of EIs. However, global attention is computationally inefficient and requires pre-training on a large-scale dataset to perform at its best. To address this issue, Panoformer \cite{panoformer} proposed pixel-based local attention for which the calculations are done by sampling nearby pixels according to the equirectangular geometry. To manage the small receptive field of local attention, Panoformer adaptively adjusts local window sizes via a learnable offset, similar to that of a deformable mechanism \cite{deformable_conv}. However, because training accurate and large offsets for a deformable mechanism is extremely difficult in practice \cite{deformable_conv2,DAT}, Panoformer is also associated with a limited receptive field.

\section{EGformer}
\label{sec:method}

\subsection{Overview} 
Figure \ref{fig:overall_architecture} illustrates the architecture of an EGformer variant (refer to Section \ref{Sec:network_variants} for more variants). Each $Horizontal_E$ (green block) and $Vertical_E$ (blue block) represents the proposed horizontal and vertical transformer blocks of EGformer (Section \ref{Sec:attention_block}), which comprises equirectangular-aware horizontal and vertical multi-head self-attention (EH\text{-}MSA,EV\text{-}MSA), as illustrated on the right side of Figure \ref{fig:overall_architecture} (Section \ref{Sec:E-MSA}). The yellow blocks (\eg, Patch embedding, Down sample) in Figure \ref{fig:overall_architecture} are based on a CNN. Details are included in the \textbf{Technical Appendix}.

\subsection{Horizontal and vertical transformer block of EGformer}
\label{Sec:attention_block}
As discussed in Section \ref{Sec:360_detph}, EIs have distinct natures along the vertical and horizontal directions \cite{360align}. The geometric structure (\eg, layout) is embedded along the vertical direction \cite{joint_360depth, dula, horizonnet}, while the cyclic structure of EIs can be addressed implicitly along the horizontal direction. For these reasons, prior studies on EIs have leveraged these natures to enhance their performance \cite{hohonet,Slicenet}. Drawing insights from these work, we adopt vertical and horizontal shaped local window for EGformer.

\begin{figure}[t!]
\centering
\begin{minipage}{.25\textwidth}
\begin{subfigure}{\linewidth}
\includegraphics[width=.98\linewidth]{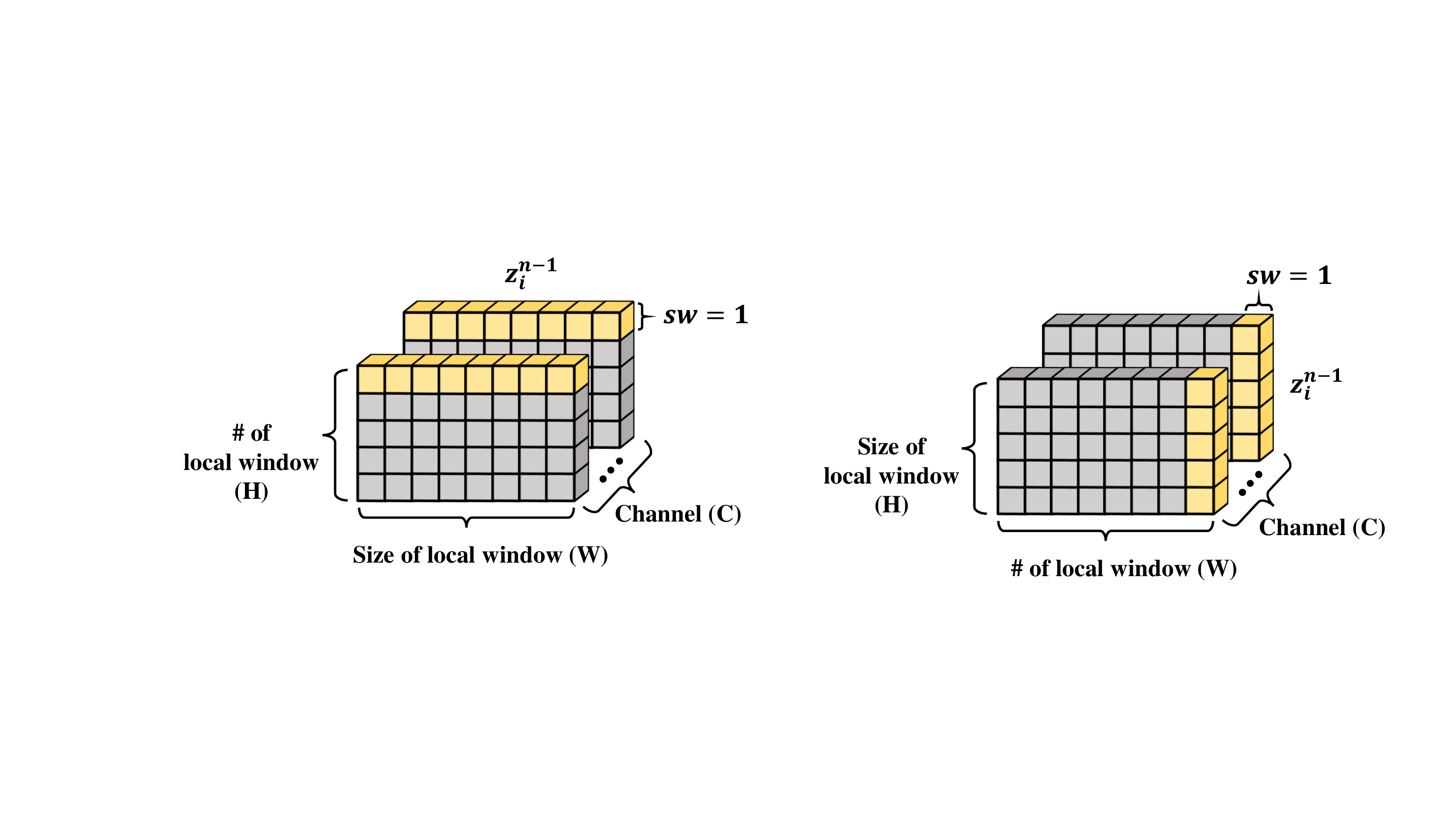}
\subcaption{horizontal local window}
\end{subfigure}
\end{minipage}%
\begin{minipage}{.25\textwidth}
\begin{subfigure}{\linewidth}
\includegraphics[width=.98\linewidth]{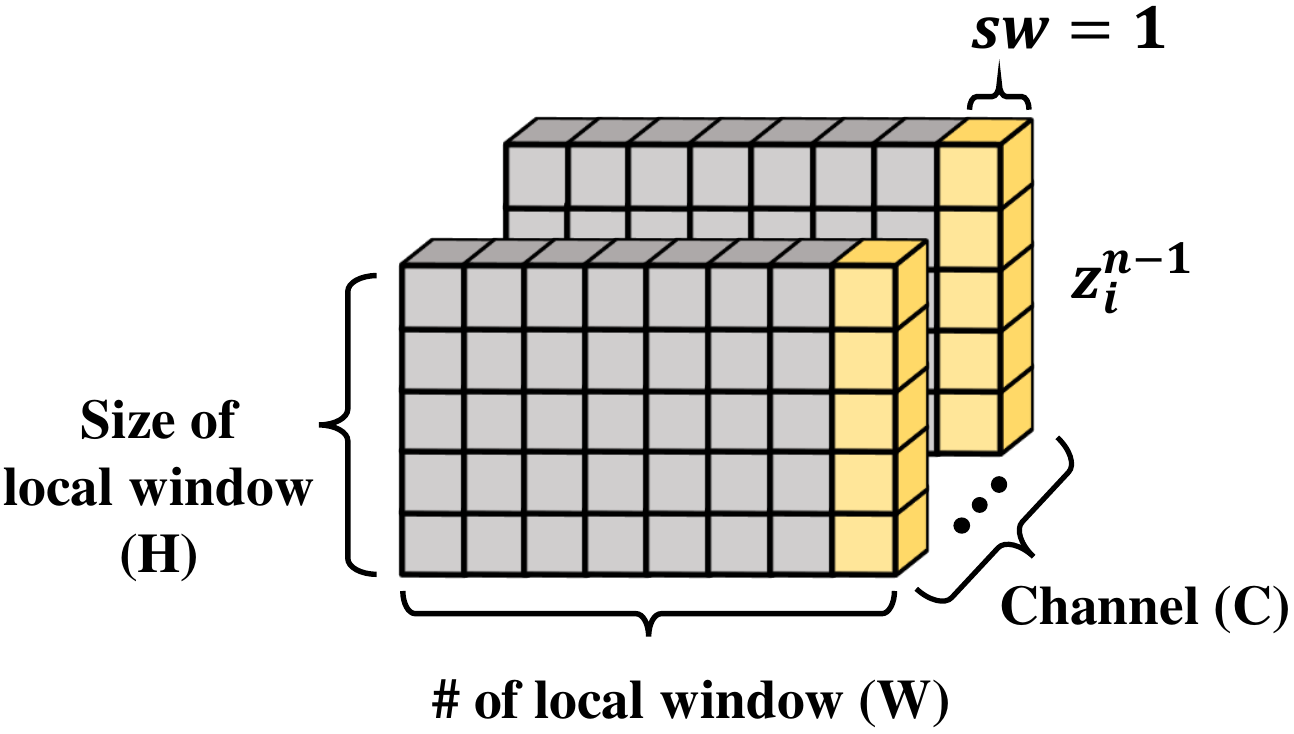}
\subcaption{vertical local window}
\end{subfigure}
\end{minipage}%
\caption{Local window shapes of EGformer}
\label{fig:local_window}
\end{figure}  

 Let us define $z^{n-1} \in \mathbb{R}^{H \times W \times (d_j \times J)}$ as the output of ($n\text{-}1$)-th transformer block or precedent convolutional layer, which are input to the $n$-th transformer block. Here, $H$ and $W$ is the height and width, where $J$ represents the number of heads in multi-head self-attention (MSA) and $d_j$ indicates the number of hidden layers of each head. In total, the channel dimension of $z^{n-1}$ is calculated via $C = d_j \times J$.
  
When $z^{n-1}$ inputs to the $Horizontal_E$, $z^{n-1}$ is divided along the horizontal direction with stripe width ($sw$) as 1 (Figure \ref{fig:local_window} (a)), constructing the group of horizontal local window features as formulated by $[z^{n-1}_1,z^{n-1}_2,\cdots,z^{n-1}_H \in \mathbb{R}^{1 \times W \times C}]$ in Eq.(\ref{Eq:EH-MSA1}). Through layer normalization ($LN$),  the normalized features of $i$-th horizontal local window for $j$-th head (\ie, $F^j_i \in \mathbb{R}^{1 \times W \times d_j}$,) is extracted via Eq.(\ref{Eq:EH-MSA2}). Then, query, key, value (\ie, $Q_i^j,K_i^j,V_i^j \in \mathbb{R}^{1 \times W \times d_j} $) are obtained by linearly projecting the $F^j_i$ as described in Eq.(\ref{Eq:EH-MSA3}). Afterwards, through proposed EH-MSA, the local attention for $i$-th horizontal local window $L^j_i \in \mathbb{R}^{1 \times W \times d_j}$ is extracted. By accumulating $L_i^j$ along the height and head dimension, equirectangular-aware horizontal attention  $L \in \mathbb{R}^{H \times W \times C}$ is constructed as shown in Eq.(\ref{Eq:EH-MSA4}). Finally, following the previous works \cite{SwinT,Cswin,DAT}, the output of $n$-th $Horizontal_E$ ($z_n$) is defined by Eq.(\ref{Eq:Final}).
  
\begin{equation}
\begin{gathered}
\label{Eq:EH-MSA1}
[z^{n-1}_1,z^{n-1}_2,\cdots,z^{n-1}_H] = z^{n-1} \\
\end{gathered}
\end{equation}

\begin{equation}
\begin{gathered}
\label{Eq:EH-MSA2}
[F^j_i]^{j=1,\cdots,J} = LN(z^{n-1}_i) \\
\end{gathered}
\end{equation}

\begin{equation}
\begin{gathered}
\label{Eq:EH-MSA3}
Q_i^j,K_i^j,V_i^j = Linear(F^j_i) \\
\end{gathered}
\end{equation}

\begin{equation}
\begin{gathered}
\label{Eq:EH-MSA4}
L_i^j = \text{EH\text{-}MSA}(Q_i^j,K_i^j,V_i^j)  \\
L = [L_1^j,L_2^j,\cdots,L_H^j]^{j=1,\cdots,J} 
\end{gathered}
\end{equation}

\begin{equation}
\begin{gathered}
\label{Eq:Final}
\hat{z}^n = L + z^{n-1}  \\
z^n = MLP(LN(\hat{z}^n)) + \hat{z}^n
\end{gathered}
\end{equation}

In the same vein, the final output of $n$-th $Vertical_E$ is extracted equally through EV-MSA with the only difference being that the group of vertical local window features (\ie, $[z_1^{n-1},z_2^{n-1},\cdots,z_W^{n-1} \in \mathbb{R}^{1 \times H \times C} ]$) is made by dividing $z^{n-1}$ along vertical direction (Figure \ref{fig:local_window} (b)).

\subsection{Equirectangular-aware horizontal and vertical self-attention}
\begin{figure}[h!]
\centering
\begin{minipage}{.465\textwidth}
\begin{subfigure}{\linewidth}
\includegraphics[width=.98\linewidth]{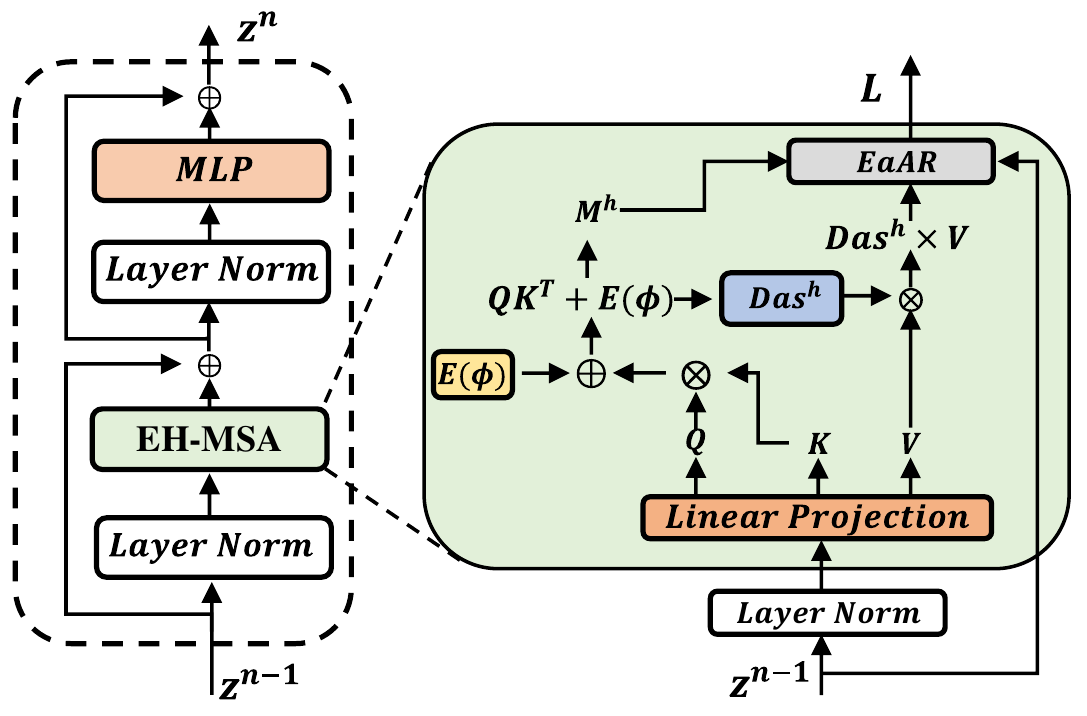}
\end{subfigure}
\end{minipage}%
\caption{Overall process of EH-MSA. The components in yellow (ERPE), blue (Das) and gray block (EaAR) denote our main proposals.}
\label{fig:EH_MSA}
\end{figure}

The overall process of EH-MSA is illustrated in Figure \ref{fig:EH_MSA}. When calculating the attention score ($QK^T + E(\phi)$), ERPE ($E(\phi)$) is added to $QK^T$ similar to relative position embedding \cite{SwinT}. Then, the $Das$ (blue block) is calculated from attention score, which produces the attention for the current block. Additionally, the importance level of each local window ($M^h$) is obtained from the attention score. Finally, through EaAR, the final attention of EH-MSA ($L$) is provided. The following subsections describe each component of E(V)H-MSA in detail.

\vspace{-10pt}

\label{Sec:E-MSA}
\paragraph{Equirectangular relative position embedding}
  We propose a non-parameterized ERPE to impose equirectangular geometry bias on the elements within each vertical and horizontal local windows. We define $E(\phi_i) \in \mathbb{R}^{W\times W}$ as the ERPE for horizontal local windows, where the ($m,n$)-th element of $E(\phi_i)$ is expressed via $E(\phi_i)_{m,n} \in \mathbb{R}^{1\times 1}$. The calculation process of $E(\phi_i)_{m,n}$ is defined by Eq.(\ref{Eq:ERPE_h}). Here, $\theta_{m,n}$ denotes the positions of the $m$-th and $n$-th elements in the horizontal local windows, where $\phi_i$ denotes the positions of the $i$-th horizontal local windows. Similarly, the ERPE for the $i$-th vertical local window is expressed via $E(\theta_i) \in \mathbb{R}^{H\times H}$, where the $(m,n)$-th element of $E(\theta_i)_{m,n} \in \mathbb{R}^{1\times 1}$ is calculated with Eq.(\ref{Eq:ERPE_v})\footnote{In experiments, we set $\rho=0.1$. Refer to Table \ref{tab:ablation_study_rho} for more details.}. The sign($\cdot$) function is used to distinguish between $E(\phi_i)_{m,n}$ and $E(\phi_i)_{n,m}$.
  
  \begin{equation}
\label{Eq:ERPE_h}
E(\phi_i)_{m,n} = \text{sign}(\theta_m -\theta_n)\cdot\rho\sqrt{2\{1 - cos(\theta_m -\theta_n) \}} \cdot sin(\phi_i)
\end{equation}
\begin{equation}
\label{Eq:ERPE_v}
E(\theta_i)_{m,n} = \text{sign}(\phi_m -\phi_n)\cdot\rho\sqrt{2\{1 - cos(\phi_m -\phi_n) \}}
\end{equation}
 
\begin{figure}[h!]
\centering
\begin{minipage}{.49\textwidth}
\begin{subfigure}{\linewidth}
\includegraphics[width=.98\linewidth]{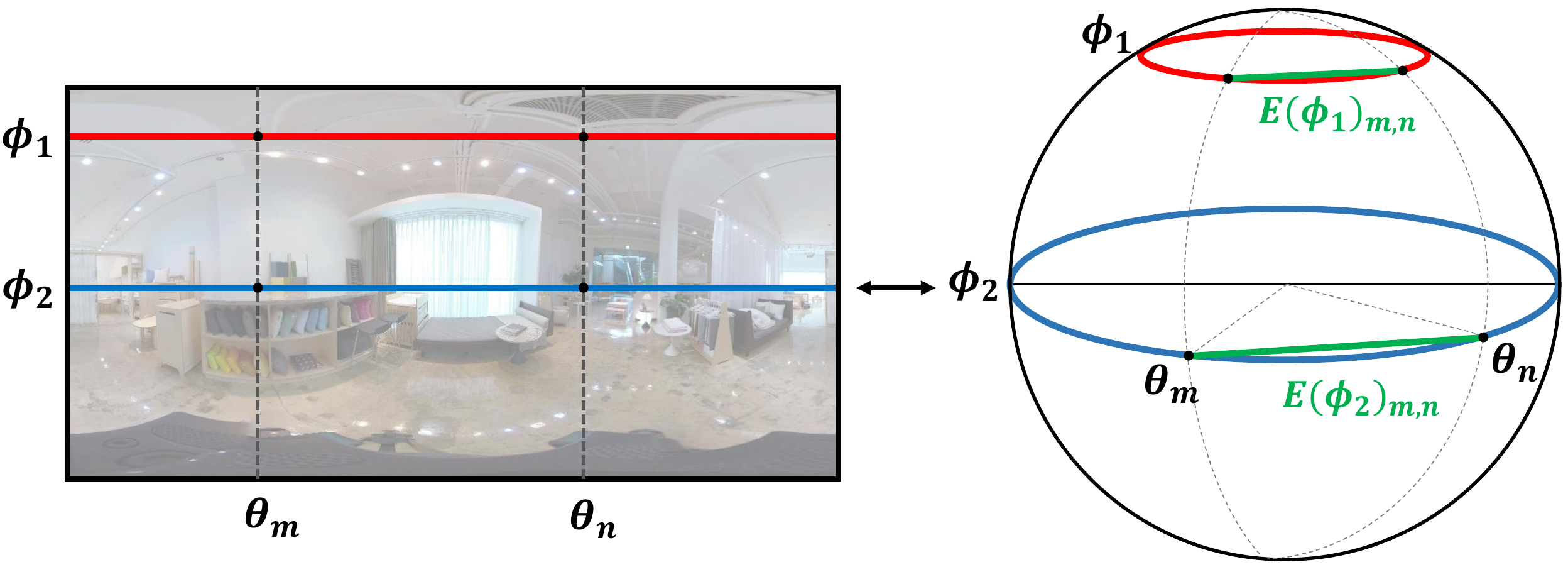}
\end{subfigure}
\end{minipage}%
\caption{ERPE for horizontal local window. Each red and blue line represents the horizontal local window.  Green line indicates the ERPE which are calculated through the distance in Cartesian coordinate. }
\label{fig:ERPE}
\end{figure}

  ERPE is calculated by measuring the distances between the $m$-th and $n$-th elements in Spherical coordinates as illustrated in Figure \ref{fig:ERPE}.
 The green line in Figure \ref{fig:ERPE} denotes $E(\phi_i)$ of each local window, while each blue and red line represents the corresponding horizontal local window.
 Unlike position embedding for general vision tasks, ERPE can enforce the attention score to assign high similarity for the elements that are close in three-dimensional space. For instance, the  ERPE of L and R in Figure \ref{fig:equirectangular_geometry} is 0 although they are far apart in EIs. As a result, ERPE induces the transformer to assign similar attention score for L and R that makes transformer to understand the cyclic structure of EIs.

\paragraph{Distance-based attention score}

Conventionally, softmax has been preferred for re-weighting the attention score. However, softmax is computationally inefficient \cite{sample_softmax1,sample_softmax2,sample_softmax3,linear_softmax,cosformer} and lacks relevance to the natural language processing (NLP) or vision tasks. For these reasons, some studies have attempted to replace softmax with alternative functions \cite{linear_softmax,cosformer}. These studies empirically determined that the essential roles performed by softmax are as follows. 

\begin{itemize}
\item{Ensuring attention score to get non-negative value}
\item{Re-weighting the attention score}
\end{itemize}

Inspired by their findings, we propose a distance-based attention score for each $i$-th horizontal ($Das^h_i \in \mathbb{R}^{1 \times W \times W}$) and vertical ($Das^v_i \in \mathbb{R}^{1 \times H \times H}$) local window, defined by Eqs.(\ref{Eq:SD_attention_h}) and (\ref{Eq:SD_attention_v}) respectively. Here, $\mathcal{N}$ represents $L_1$ normalization, and $(\rho_b,\theta_b,\phi_b)$ represents the baseline point (hyperparameter). Then, the local attention of each $i$-th local window is obtained using Eq.(\ref{Eq:Attention})

\begin{equation}
\begin{gathered}
\label{Eq:Attention_score}
{score}_i^h = Q_iK_i^T + E(\phi_i) \\
{score}_i^v = Q_iK_i^T + E(\theta_i) \\
\end{gathered}
\end{equation}

\begin{equation}
\label{Eq:SD_attention_h}
Das^{h}_i = 2{\rho_b}^2\cdot\{1 - cos(\mathcal{N}\{{score}_i^h\}\cdot{\pi \over 2})\}\cdot sin^2(\phi_b)
\end{equation}

\begin{equation}
\label{Eq:SD_attention_v}
Das^{v}_i = 2{\rho_b}^2\cdot\{1 - cos(\mathcal{N}\{{score}_i^v\}\cdot {\pi \over 2})\}
\end{equation}

\begin{equation}
\begin{gathered}
\label{Eq:Attention}
Attention_i^{h} = Das_i^{h} \times V_i \\
Attention_i^{v} = Das_i^{v} \times V_i
\end{gathered}
\end{equation}

\begin{figure}[h!]
\centering
\begin{minipage}{.49\textwidth}
\begin{subfigure}{\linewidth}
\includegraphics[width=.98\linewidth]{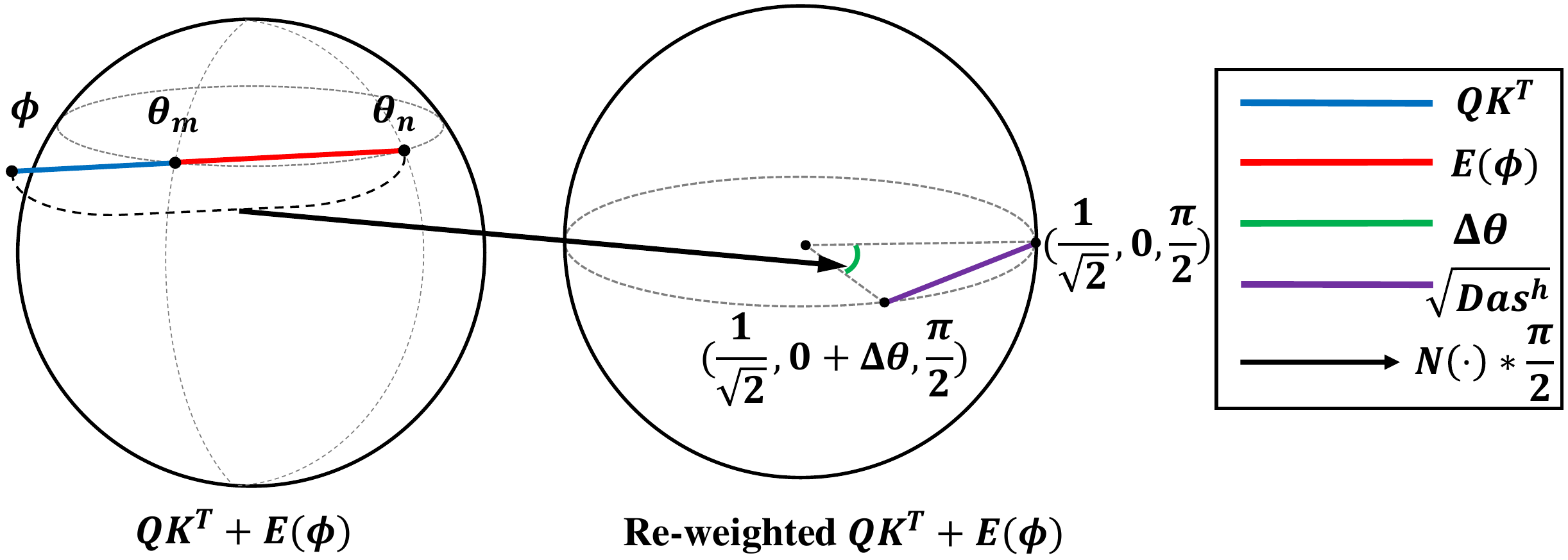}
\end{subfigure}
\end{minipage}%
\caption{The illustration of the distance-based attention score for horizontal local window. Each element in $score^h_i$ is re-weighted to the distances from the $({1\over \sqrt{2}},0,{\pi\over 2})$, which determines the distance-based attention score of each element.}
\label{fig:Das}
\end{figure}

 The core idea of $Das_i^{h,v}$ is to convert each element of ${score}_i^{h,v}$ into the distances from the baseline point $(\rho_b,\theta_b,\phi_b)$ in Spherical coordinates. Simply put, the farther the element of ${score}_i^{h,v}$ is from $(\rho_b,\theta_b,\phi_b)$, the higher the distance-based attention score it receives. In this paper, we set baseline point as $(\rho_b,\theta_b,\phi_b)$=$({1\over \sqrt{2}},0,{\pi\over 2})$ to make both $Das^{h}$ and $Das^{v}$ get equal score range $[0,1]$. The calculation process of $Das^h_i$ is visualized in Figure \ref{fig:Das}, which is performed via the following steps.  First, through normalization as denoted by the black arrow in Figure \ref{fig:Das}, ${score}_i^{h}$ is converted to $\Delta \theta \in (-{\pi \over 2},{\pi \over 2})$, as visualized by the green curve in Figure \ref{fig:Das}. 
Second, by calculating the distance of $({1\over \sqrt{2}},0 + \Delta \theta,{\pi \over 2})$ from $({1\over \sqrt{2}},0,{\pi\over 2})$ in Spherical coordinates, $\sqrt{Das^h_i}$ is obtained, as represented via the purple line in Figure \ref{fig:Das}. Finally, as shown in Eq.(\ref{Eq:SD_attention_h}), square of the distance is calculated for the final distance-based attention score to focus more on the important region by amplifying the differences in score value. In the same vein, ${score}_i^{v}$ is converted to $\Delta \phi \in (-{\pi \over 2},{\pi \over 2}) $ . Then, by calculating the square of the distance of $({1\over \sqrt{2}}, 0,  {\pi \over 2} + \Delta \phi)$ from $({1\over \sqrt{2}},0,{\pi\over 2})$ in Spherical coordinates, $Das^v_i$ is obtained as described in Eq.(\ref{Eq:SD_attention_v}).

Compared to softmax which is biased in positive values\footnote{To get high softmax attention score, an element in $score$ should have higher 'positive' values than others.}, $Das$ is more appropriate for ERPE. Because $E(\theta,\phi)_{m,n}$=$-E(\theta,\phi)_{n,m}$, softmax forces unbalanced $score$ for $(m,n)$-th and $(n,m)$-th elements when ERPE is imposed. Unlike the NLP in which the order of each element matters, it is often not for depth estimation tasks. Therefore, unbalanced $score$ may confuse a transformer, resulting in incorrect score extraction. On the contrary, $Das$ is symmetric \footnote{ 1-$cos(x)$=1-$cos(-x)$}. Therefore, when ERPE is used with $Das$, a transformer can get balanced $score$ if required, which would be more suitable for EI depth estimation.

\paragraph{Equirectangular-aware attention rearrangement}
Although structural prior of EIs is embedded in $Attention_i$ via ERPE and $Das$, structural prior between $Attention_i$ is not yet imposed. To address this issue, we propose equirectangular-aware attention rearrangement, defined by Eq.(\ref{Eq:Mixed_Attention}). $L_i^h \in \mathbb{R}^{1 \times W \times C} $ and $L_i^v \in \mathbb{R}^{1 \times H \times C} $ indicate the rearranged local attention of the $i$-th horizontal and vertical local window respectively, with $M^h_i \in \mathbb{R}^{1 \times 1 \times 1}$ and $M^v_i \in \mathbb{R}^{1\times 1 \times 1}$ representing the importance level of each local window. The importance level of each local window is defined by Eq.(\ref{Eq:Mixed_weight_h}), where $f(x)$ calculates the mean of all elements in $x$. Local windows that are important receive $M$ values close to 1, while those that are unimportant receive $M$ values close to 0 \footnote{In experiments, we clamp $M_i$ to have value of 0.5 at its minimum to ensure certain amount of $Attention_i$ to be used for $L$}.

\begin{equation}
\begin{gathered}
\label{Eq:Mixed_Attention}
L_i^h = M^{h}_i \cdot (Attention_i) + (1 - M^{h}_i) \cdot z^{n-1}_i \\
L_i^v = M^{v}_i \cdot (Attention_i) + (1 - M^{v}_i) \cdot z^{n-1}_i
\end{gathered}
\end{equation}

\begin{equation}
\label{Eq:mean_function}
f(x) = {\sum_{p=1}^N{x_p} \over N}
\end{equation}

\begin{equation}
\label{Eq:Mixed_weight_h}
M^{h}_i = {{f(|{score}_i^h|) \over \underset{\forall i}{\max}({f(|{score}_i^h|)})}}, M^{v}_i = {{f(|{score}_i^v|) \over \underset{\forall i}{\max}({f(|{score}_i^v|)})}}
\end{equation}

To approximate $M_i^{h,v}$, we utilize $score_i^{h,v}$. Because $score_i^{h,v}$ is equirectangular geometry-biased, the mean of $|score_i^{h,v}|$ implicitly reflects both the information density and distinctive features of each local window. 
 Specifically, $E(\theta,\phi)$ term in $score_i^{h,v}$ is closely related to the information density\footnote{$f(|E(\theta_i,\phi_i)|)$ gets proportional relationship with the information density of each $i$-th local window as shown in Figure \ref{fig:ERPE}.} and $QK^T$ term is related to the distinctive characteristics of each local window.  Because the information density of local window have high relevance with the level of importance, geometry-biased score values (\ie, $|score_i^{h,v}|$) are appropriate to estimate $M^{h,v}$. 

 As shown in Eq.(\ref{Eq:Mixed_weight_h}), the final importance level of $i$-th local window is obtained by comparing the $f(|score_i^{h,v}|)$ of all other local windows. Therefore, $M^{v,h}_i * Attention^{h,v}_i$ term achieve global-like characteristics by making the local attention to interact with each other indirectly. 
 As a result, the local attention can be extracted more accurately from a feature map with a high resolution. This enables the retention of detailed spatial information and ultimately improves the depth quality.
 However, in practice, the importance level of each local window can be predicted incorrectly. This could potentially dilute the important information of $Attention^{h,v}_i$ via multiplication with $M^{v,h}_i$. The $(1-M^{v,h}_i) * z^{n-1}_i$ term can prevent such a situation and enable $L_i$ to be extracted by observing various attention blocks simultaneously, resulting in a more global representation of $L_i$.

\paragraph{Computational complexity}
The computational complexity of EH(V)-MSA is as follows: 
\begin{equation}
\begin{gathered}
\label{Eq:computational_complexity}
\Omega(\text{EH-MSA}) = 4HWC^2 + 2HW^2C \\
\Omega(\text{EV-MSA}) = 4HWC^2 + 2H^2WC
\end{gathered}
\end{equation}

\section{Experiments}
\label{sec:experiments}
Due to page limitation, detailed experimental environment is described in the Technical Appendix.

\subsection{Experimental environment}
\label{Sec:experiment_environment}

\noindent \paragraph{\noindent \textbf{Dataset}} \noindent
We evaluate our method using Structured3D \cite{structure3d} and Pano3D \cite{Pano3D} datasets, which are the most recent datasets with the highest quality. Discussions on other datasets \cite{matterport,stanford,omnidepth,360self} are included in Technical Appendix. 

\noindent \paragraph{\noindent\textbf{Metrics}} \noindent
 The scale of the depth differs according to how the depth is acquired; therefore, an alignment process is commonly used when evaluating the depths of multiple dataset simultaneously \cite{silog_loss,ordinal_loss,NMG_loss,midas,DPT,joint_360depth}.   Following earlier work \cite{DPT}, we align the depths in an image-wise manner before measuring the errors for each dataset as defined by Eq.(\ref{Eq:align}) for all methods. Quantitative results are extracted by comparing aligned depth ($Depth_A$) with the ground truth ($GT$). 
 
\begin{equation}
\begin{gathered}
\label{Eq:align}
s,t = \argmin_{s,t}(s \cdot {Depth} + t - GT) \\
Depth_A = s\cdot Depth + t
\end{gathered}
\end{equation}

Common evaluation metrics are used. Lower is better for the absolute relative error (Abs.rel), squared relative error (Sq.rel), root mean square linear error (RMS.lin), root mean square log error (RMSlog). Meanwhile, higher is better for relative accuracy ($\delta^n$), where $\delta^n$ represents $\delta <1.25^n$. FLOPs are calculated using Structured3D testset \cite{structure3d}.

\begin{table*}[t!]
\renewcommand{\tabcolsep}{1mm}
\small
\centering
\begin{tabular}{c|c|c|cccc|ccc|c|c}
\hline
ID&Encoder & Decoder & Abs.rel & Sq.rel & RMS.lin  & RMSlog & $\delta^1$ & $\delta^2$ & $\delta^3$ &$\#$Param &FLOPs\\ \hline
0& HHHH &  HHHH & 0.0421 & 0.0373 & 0.2961 &0.1007 &0.9784 &0.9922 & 0.9960 &15.3M & 81.5G \\
1& VVVV &  VVVV & 0.0389 & 0.0346 & 0.2983 & 0.0998 & 0.9782 & 0.9920 & 0.9959 & 15.3M & 70.1G \\
2& EEEE &  EEEE  &  0.0375 & 0.0320 & 0.2945 & 0.0979 &  0.9782 & 0.9920 & 0.9960 & 15.3M & 75.8G \\
3& MMMM &  MMMM &  0.0366 & 0.0308 & 0.2795 &0.0959 & 0.9798 & 0.9926 &0.9963 & 17.5M & 80.1G  \\
4& PPEE &  EEPP & 0.0362 & 0.0318  &0.2874 & 0.0979 &  0.9791 & 0.9921 & 0.9960 & 15.6M & 77.6G \\
5& MMEE &  EEMM  &  0.0342  & 0.0279  & 0.2756 & 0.0932 &  0.9810 & 0.9928 & 0.9964 & 15.4M & 73.9G \\
\hline

\end{tabular}
\caption{Depth estimation results on network variants for Structured3D testset \cite{structure3d}. Bottleneck layer is fixed to \textbf{E}.} 
\label{tab:network_architecture}       
\end{table*} 
\begin{table*}[t!]
\renewcommand{\tabcolsep}{1mm}
\small
\centering

\begin{tabular}{ccc||cccc|ccc|c|c}
\hline
Testset & Method & Backbone & Abs.rel & Sq.rel & RMS.lin & RMSlog & $\delta^1$ & $\delta^2 $ & $\delta^3$ & $\#$Param & FLOPs \\ \hline

\multirow{5}{*}{Structured3D \cite{structure3d}} & Bifuse \cite{bifuse} & CNN & 0.0644 & 0.0565 & 0.4099 & 0.1194 & 0.9673 & 0.9892 & 0.9948 & 253.0M & 723.4G \\
& SliceNet \cite{Slicenet} &CNN+RNN & 0.1103 & 0.1273 & 0.6164 & 0.1811 & 0.9012 & 0.9705 & 0.9867 &79.5M& 84.3G \\
& Yun \etal \cite{joint_360depth} & Global & 0.0505 & 0.0499 & 0.3475 & 0.1150 & 0.9700 & 0.9896 & 0.9947 & 123.7M & 589.4G \\
& Panoformer \cite{panoformer} & Local & 0.0394 & 0.0346 & 0.2960 & 0.1004 &0.9781  & 0.9918 & 0.9958 & 20.4M & 77.7G \\ 
& EGformer & Local & \bfseries 0.0342 &\bfseries 0.0279 &\bfseries 0.2756 &\bfseries 0.0932 &\bfseries 0.9810 &\bfseries 0.9928 &\bfseries 0.9964 &\bfseries 15.4M & \bfseries 73.9G  \\ 
\hline

\multirow{5}{*}{Pano3D \cite{Pano3D}} & Bifuse \cite{bifuse} &CNN & 0.1704 & 0.1528 & 0.7272 & 0.2466 & 0.7680 & 0.9251 & 0.9731 & 253.0M & 723.4G\\
& SliceNet \cite{Slicenet} &CNN+RNN & 0.1254 & 0.1035 & 0.5761 & 0.1898 & 0.8575 & 0.9640 & 0.9867 & 79.5M & 84.3G \\
& Yun \etal \cite{joint_360depth} & Global & 0.0907 & 0.0658 & 0.4701 & 0.1502 & 0.9131 & 0.9792 & 0.9924 &123.7M & 589.4G \\
& Panoformer \cite{panoformer} & Local & 0.0699 & 0.0494 & 0.4046 & 0.1282 & 0.9436  & 0.9847 & 0.9939 & 20.4M & 77.7G \\ 
& EGformer & Local &\bfseries 0.0660 &\bfseries 0.0428 &\bfseries 0.3874 &\bfseries 0.1194 &\bfseries 0.9503 &\bfseries 0.9877 &\bfseries 0.9952 & \bfseries 15.4M & \bfseries 73.9G \\ 
\hline

\end{tabular}
\caption{Quantitative depth results of each method. Numbers in \textbf{bold} indicate the best results. It is observed that EGformer achieves the best depth outcomes with the lowest FLOPs and the fewest parameters. }
\label{tab:retrained_depth}       
\end{table*} 
\subsection{Model study}
\label{Sec:network_variants}
\begin{figure}[h!]
\centering
\begin{minipage}{.15\textwidth}
\begin{subfigure}{\linewidth}
\includegraphics[width=.98\linewidth]{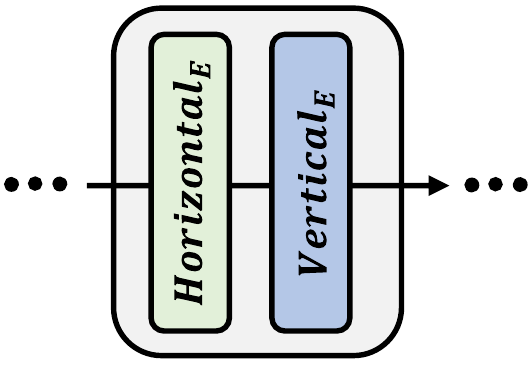}
\subcaption{Module \textbf{E}}
\end{subfigure}
\end{minipage}%
\begin{minipage}{.15\textwidth}
\begin{subfigure}{\linewidth}
\includegraphics[width=.98\linewidth]{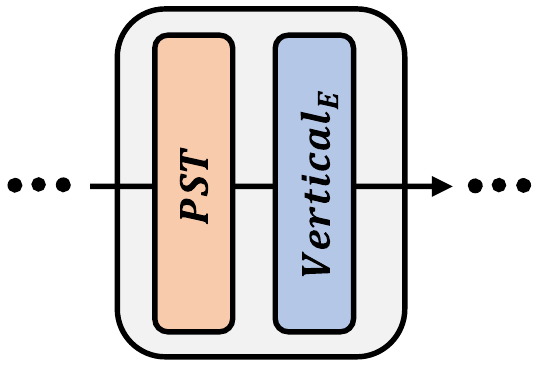}
\subcaption{Module \textbf{M}}
\end{subfigure}
\end{minipage}%
\begin{minipage}{.15\textwidth}
\begin{subfigure}{\linewidth}
\includegraphics[width=.98\linewidth]{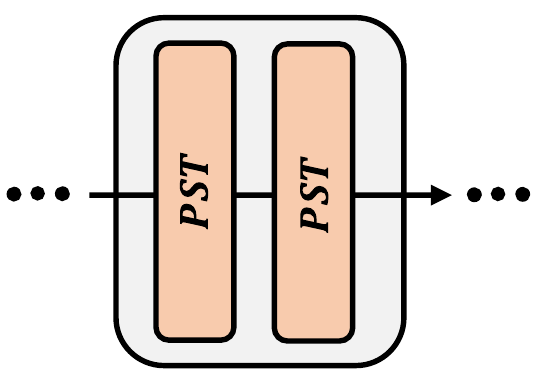}
\subcaption{Module \textbf{P}}
\end{subfigure}
\end{minipage}%

\begin{minipage}{.15\textwidth}
\begin{subfigure}{\linewidth}
\includegraphics[width=.98\linewidth]{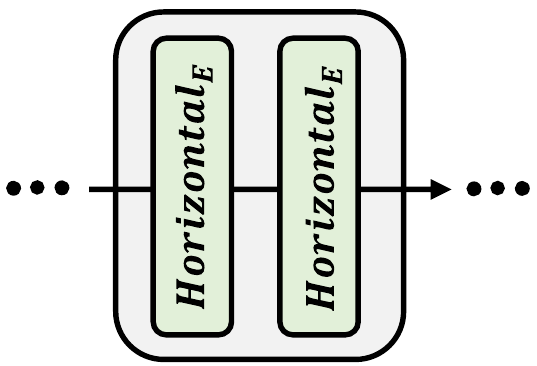}
\subcaption{Module \textbf{H}}
\end{subfigure}
\end{minipage}%
\begin{minipage}{.15\textwidth}
\begin{subfigure}{\linewidth}
\includegraphics[width=.98\linewidth]{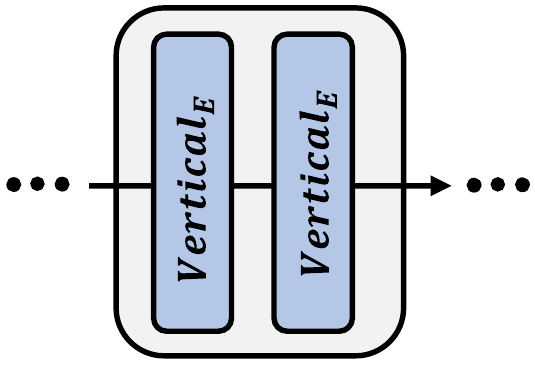}
\subcaption{Module \textbf{V}}
\end{subfigure}
\end{minipage}%
\caption{Various attention module used in Table \ref{tab:network_architecture}. $PST$ represents the attention block of Panoformer \cite{panoformer}.}
\label{fig:network_architecture}
\end{figure}

 Because all methods have their pros and cons, it is often observed that combining several methods yields better outcomes. For this reason, we study various EGformer variants. Figure \ref{fig:network_architecture} shows the various attention module as denoted by \textbf{E, M, P, H} and  \textbf{V}. Here, $PST$ indicates Panoformer attention block \cite{panoformer}. Based on this annotation, network architecture can be expressed simply. For example, network architecture in Figure \ref{fig:overall_architecture} is expressed via 'EEEE-E-EEEE'. 
 
Table \ref{tab:network_architecture} shows the quantitative results of each network variant for the Structured3D testset \cite{structure3d}. As shown in Table \ref{tab:network_architecture}, the \textbf{H} or \textbf{V} attention module does not yield plausible depth results as shown in ID 0 and 1. As discussed in previous studies \cite{Cswin}, the relatively poor performances of \textbf{H}/\textbf{V} modules can be explained via their narrow stripe widths (\ie, $sw=1$). Although consecutive horizontal and vertical attention module (\textbf{E}) can alleviate the problem of a narrow stripe width, as shown in ID 2, this solution falls short. Under the circumstances, the easiest means of improving depth quality level is to enlarge the stripe width \cite{Cswin}. However, a wider stripe width also increases the computational cost significantly. Therefore, instead of using wider stripes, we attempt to improve the depth quality by mixing various attention modules, as shown in ID 3, 4 and 5. Among these, we observe that ID 5 is the best fit for our purpose. Based on these results, we set the network architecture of ID 5 as the default architecture in this paper. Meanwhile, the performance differences between ID 3,4 and ID 5 clearly demonstrate the effect of the proposed EH(V)-MSA.

 \begin{figure*}[t!]
\centering
\begin{minipage}{.35\textwidth}
\begin{subfigure}{\linewidth}
\includegraphics[width=.98\linewidth]{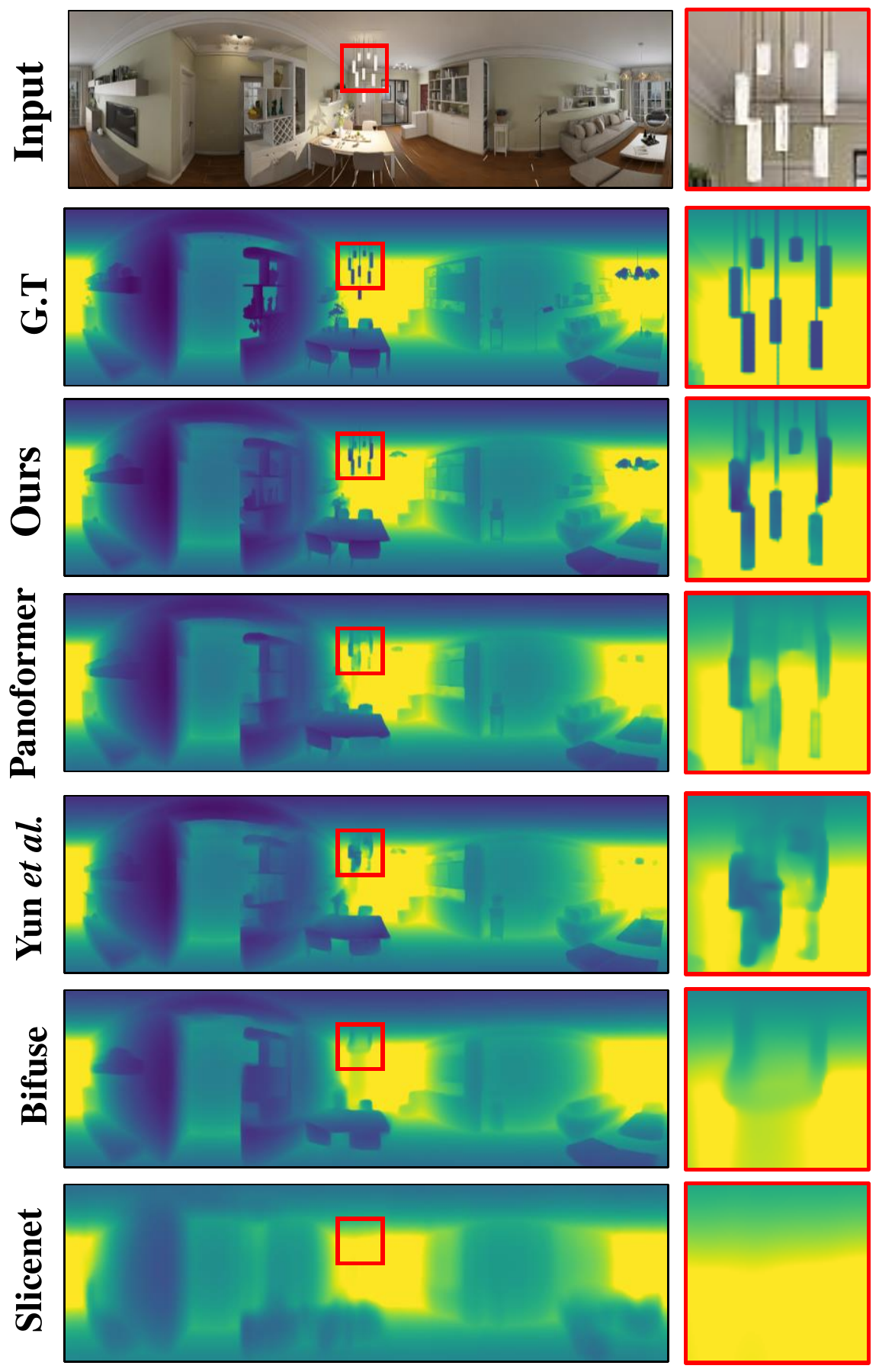}
\subcaption{Sturctured3D\cite{structure3d}}
\end{subfigure}
\end{minipage}%
\begin{minipage}{.35\textwidth}
\begin{subfigure}{\linewidth}
\includegraphics[width=.98\linewidth]{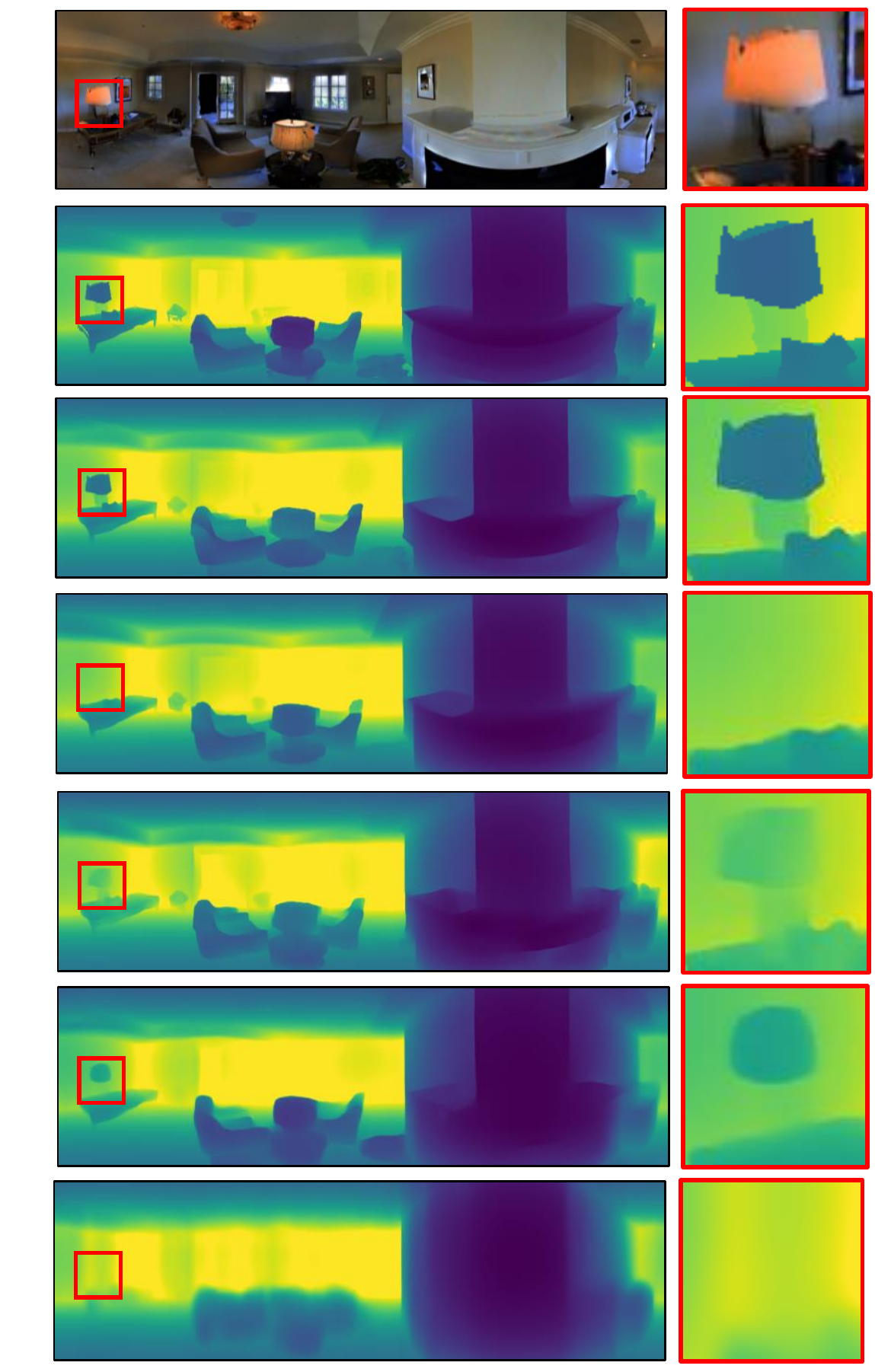}
\subcaption{Pano3D\cite{Pano3D}}
\end{subfigure}
\end{minipage}%
\begin{minipage}{.337\textwidth}
\begin{subfigure}{\linewidth}
\includegraphics[width=.98\linewidth]{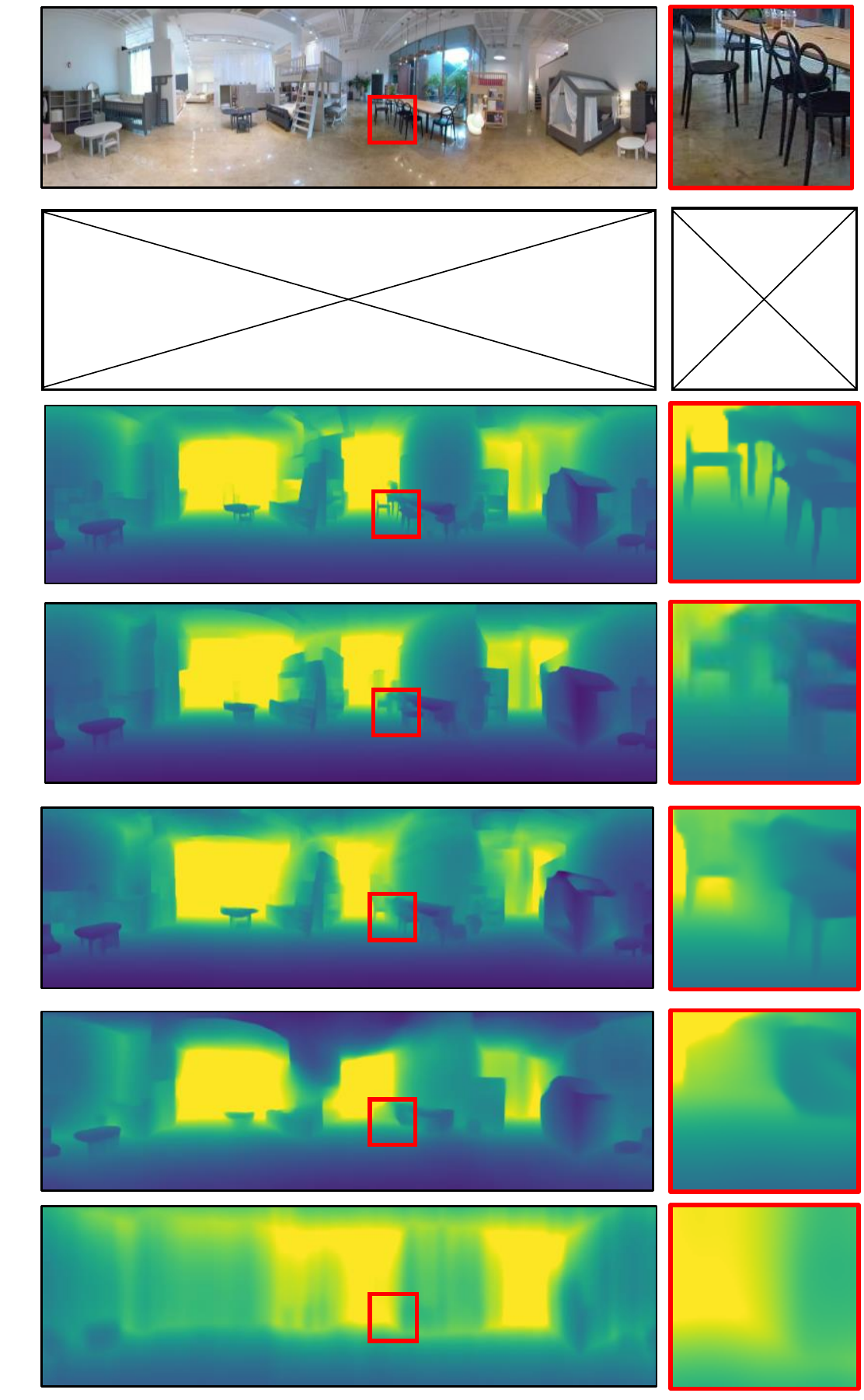}
\subcaption{Real-world scene\cite{joint_360depth}}
\end{subfigure}
\end{minipage}%
\caption{Qualitative results of each method. G.T represents ground truth. 
 Due to page limitations, top-down parts of images are cropped in this figure. There are no ground truth depths in (c) Real-world scene \cite{joint_360depth}.}
\label{fig:qualitative_results}
\end{figure*}

\subsection{Comparison with state-of-the-arts}

To demonstrate the effectiveness of our proposals, we compare EGformer with the state-of-the-arts. Table \ref{tab:retrained_depth} shows the quantitative results of each method. Compared to CNN or RNN based approaches, it is observed that transformer-based approaches yield much better depth outcomes. Among which, EGformer achieves the best depth outcomes with the lowest computational cost and the fewest parameters. The reason for low depth qualities of Yun \etal \cite{joint_360depth} could be the lack of dataset. Vision transformer, in which Yun \etal based on, requires large-scale dataset \cite{vision_transformer}. However, under our experimental environment, scarce equirectangular depth dataset is used only, which may impair the performances. A further analysis is included in Technical Appendix.

Figure \ref{fig:qualitative_results} shows the qualitative results of each method. As similar to Table \ref{tab:retrained_depth}, transformer-based approaches provide much better results than that of CNN or RNN based approaches. Slicenet \cite{Slicenet} lacks details as reported similarly by \cite{panoformer}, and Bifuse \cite{bifuse} provides unsatisfactory results considering the large computational cost and parameters. Out of transformer-based approaches, EGformer yields the best performance in terms of the details. In particular, Panoformer fails to reconstruct the depth of a lamp in Pano3D testset.  On the contrary, EGformer reconstructs the depth of a lamp successfully. This result supports our argument in that EGformer extracts the attention accurately even from a feature map with a high resolution, which enables to keep the detailed spatial information.
The results on challenging real-world scenes further demonstrate our arguments. All methods fail to distinguish chairs from the background except the EGformer.

\subsection{Ablation study}
As discussed in Section \ref{Sec:E-MSA}, each component of EH(V)-MSA is engineered to perform at its best when they are used altogether. For example, ERPE requires symmetric characteristics of $Das$ to impose geometry bias naturally, and EaAR requires a well-biased $score$ to rearrange the attention properly. Table \ref{tab:ablation_study} shows these characteristics. Here, ID 4 uses softmax attention score with locally enhanced position encoding (LePE) \cite{Cswin}. Compared to ID 0, a significant performance drop is observed when each component of EH(V)-MSA is removed from ID 0 as shown in ID 1,2,3 and 4. Although these dependencies can be seen as a weakness of EH(V)-MSA, they also suggest that EH(V)-MSA is elaborately designed, which explains why EGformer enables the efficient extraction of the attention for EIs.

\begin{table}[t!]
\renewcommand{\tabcolsep}{1mm}
\scriptsize
\centering
\begin{tabular}{c|c|ccc||cccc|c}
\hline
Data & ID & ERPE & Das & EaAR & Abs.rel & Sq.rel & RMSlin & RMSlog  &  $\delta^1 $ \\ \hline
\multirow{5}{*}{S3D}& 0 & \checkmark & \checkmark & \checkmark & \bfseries 0.0342  & \bfseries 0.0279  &\bfseries 0.2756&\bfseries  0.0932 &\bfseries  0.9810   \\
& 1 &  \checkmark &  & \checkmark & 0.0360 &  0.0307   & 0.2804 & 0.0948  &  0.9799  \\
& 2 &  \checkmark & \checkmark & & 0.0363  &  0.0301  & 0.2804 & 0.0966  &  0.9805  \\
& 3 & \checkmark & && 0.0374  &  0.0318  & 0.2914 & 0.0984  &  0.9791  \\
& 4 &  & &  &   0.0371  &  0.0316 & 0.2859 & 0.0971  & 0.9793 \\
\hline

\multirow{5}{*}{Pano3D}& 0 &\checkmark &\checkmark & \checkmark & \bfseries 0.0660  &\bfseries  0.0428  & \bfseries 0.3874 &\bfseries 0.1194 &\bfseries 0.9503   \\
&1 &\checkmark  && \checkmark & 0.0677   & 0.0443 & 0.3972 &0.1225  &  0.9479  \\
&2 & \checkmark &\checkmark & & 0.0687  &  0.0449  & 0.3966 &0.1228  & 0.9479  \\
&3 & \checkmark & & & 0.0689   & 0.0448   & 0.3983 &0.1227  & 0.9482   \\
& 4 & & &  & 0.0700  &  0.0466 & 0.4052 &0.1248 & 0.9455   \\
\hline

\end{tabular}
\caption{Ablation study. ID 4 uses softmax attention score with locally enhanced position encoding (LePE) \cite{Cswin}. S3D represents the Structured3D testset \cite{structure3d}. }
\label{tab:ablation_study}      
\end{table}

\begin{figure}[h!]
\centering
\begin{minipage}{.47\textwidth}
\begin{subfigure}{\linewidth}
\includegraphics[width=.98\linewidth]{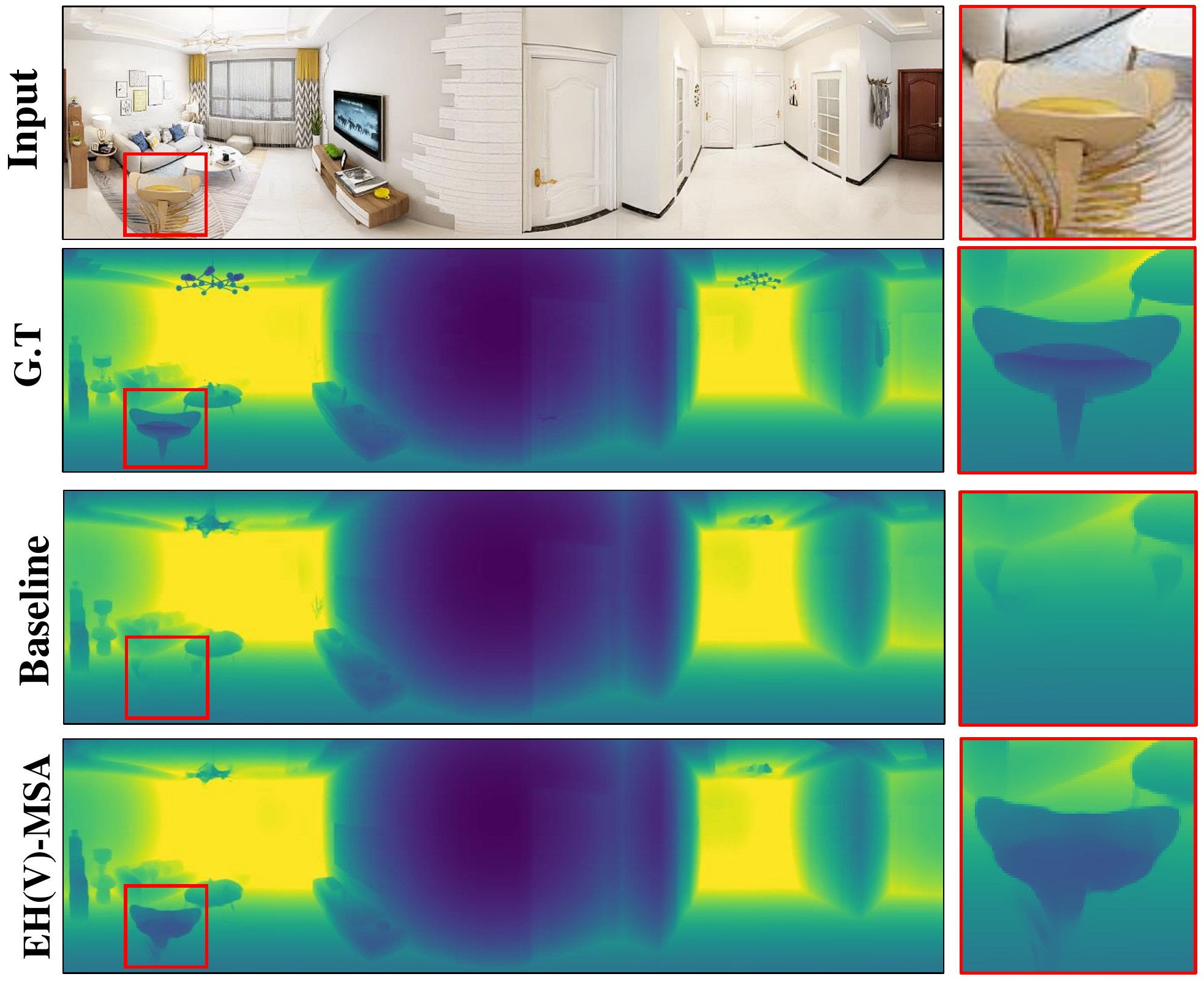}
\end{subfigure}
\end{minipage}%
\caption{Effect of EH(V)-MSA for 'EEEE-E-EEEE' network architecture. Baseline uses softmax attention score with LePE \cite{Cswin}.}
\label{fig:ablation_study}
\end{figure}

\paragraph{Further study on the effect of EH(V)-MSA} Because the network architecture of EGformer in Table \ref{tab:ablation_study} is '\textbf{MMEE-E-EEMM}', $PST$ in \textbf{M} may dilute the effect of EH(V)-MSA.  Therefore, to see the effect of EH(V)-MSA more clearly, we conduct an additional ablation study. Table \ref{tab:ablation_study2} and Figure \ref{fig:ablation_study} show the depth estimation results of '\textbf{EEEE-E-EEEE}' architecture for Structured3D testset.  As equal to ID 4 in Table \ref{tab:ablation_study}, 'Baseline' represents the model that uses softmax attention score with LePE \cite{Cswin} as similar to that of CSwin transformer \cite{Cswin}. As shown in Table \ref{tab:ablation_study2}, improvements are observed when EH(V)-MSA is used instead of CSwin attention mechanism. Meanwhile, Figure \ref{fig:ablation_study} shows interesting results. As similar to the result of Panoformer in Figure \ref{fig:qualitative_results} (b), it is shown that Baseline model fails to reconstruct the depth of a small chair as shown in Figure \ref{fig:ablation_study}.  These results further support our arguments in that the lack of details in depths are common limitation of small receptive field regardless of the shape of the local window.  On the contrary, EH(V)-MSA reconstructs the depth of a chair appropriately. This demonstrates clearly in that EH(V)-MSA acts as a key role in keeping the detailed spatial information. 

\begin{table}[h!]
\renewcommand{\tabcolsep}{1mm}
\scriptsize
\centering
\begin{tabular}{c|cccc|ccc}
\hline
 Method &  Abs.rel & Sq.rel & RMS.lin & RMSlog  &  $\delta^1 $ &  $\delta^2 $ &  $\delta^3 $ \\ \hline
 Baseline  &   0.0399  &  0.0358 &  0.3016  & 0.1014 & 0.9766 & 0.9916 & 0.9958 \\
 EH(V)-MSA & 0.0375 & 0.0320 & 0.2945 &  0.0979 &  0.9782  & 0.9920  &  0.9960  \\

\hline

\end{tabular}	
\caption{Depth estimation results when different attention mechanism is used for 'EEEE-E-EEEE' architecture for Structured3D testset \cite{structure3d}. Baseline uses softmax attention score with LePE \cite{Cswin}. }
\label{tab:ablation_study2}      
\end{table}  

\paragraph{Ablation study on bias level ($\rho$)} Although it is demonstrated that equirectangular geometry bias is effective in extracting accurate local attention, excessive geometry bias also may dilute the important information in attention. In EGformer, the influence of equirectangular geometry bias on attention is controlled by $\rho$ in Eqs.(\ref{Eq:ERPE_h}) and (\ref{Eq:ERPE_v}). Therefore, to find the appropriate bias level ($\rho$), we conduct an experiment in Table \ref{tab:ablation_study_rho}, which shows the depth estimation results when different bias level is used. Unlike previous experiments, Pano3D dataset is not used here for training. As shown in Table \ref{tab:ablation_study_rho}, the best results are observed when $\rho=0.1$. These results show that appropriate bias level is important for accurate attention. Meanwhile, based on the results in Table \ref{tab:ablation_study_rho}, we used $\rho=0.1$ as default in this paper.

\begin{table}[h!]
\renewcommand{\tabcolsep}{1mm}
\scriptsize
\centering
\begin{tabular}{c|cccc|ccc}
\hline
 $\rho$ &  Abs.rel & Sq.rel & RMS.lin & RMSlog  &  $\delta^1 $ &  $\delta^2 $ &  $\delta^3 $ \\ \hline
 0.03 & 0.0347 & 0.0284 & 0.2765 &  0.0941 &  0.9811  &\bfseries 0.9930  &  0.9963  \\
 0.1  &  \bfseries 0.0338  &\bfseries  0.0268 &\bfseries  0.2731  &\bfseries 0.0933 &\bfseries 0.9816 & 0.9929 & 0.9963 \\
 0.3 & 0.0352 & 0.0288 & 0.2747 &  0.0942 &  0.9811  & 0.9924  &\bfseries  0.9964  \\

\hline

\end{tabular}	
\caption{Depth estimation results when different bias level ($\rho$) is used for 'MMEE-E-EEMM' architecture. Each model is trained and tested with Structured3D dataset \cite{structure3d}. }
\label{tab:ablation_study_rho}      
\end{table}

\section{Conclusion}
\label{sec:conclusion}

In this paper, we propose EGformer for efficient and sophisticated equirectangular depth estimation. The core of EGformer is E(H)V-MSA, which enables to extract local attention in a global manner by considering the equirectangular geometry. To achieve this, we actively utilize the structural prior of EIs when extracting the local attention. Through experiments, we demonstrate that EGformer enables to improve the depth quality level while limiting the computational cost. Considering that EGformer can be generally applied with other attention block as demonstrated in experiments, we expect that EGformer will be extremely beneficial for various 360 vision tasks. 

\section*{Acknowledgments}
This work was supported by Samsung Research Funding Center of Samsung Electronics under Project Number SRFC-IT1702-54 and by Institute of Information $\&$ communications Technology Planning $\&$ Evaluation (IITP) grant funded by the Korea government(MSIT) (No.RS-2022-00155915, Artificial Intelligence Convergence Innovation Human Resources Development (Inha University)).

{\small
\bibliographystyle{ieee_fullname}
\bibliography{egbib}

\begin{thebibliography}{10}\itemsep=-1pt

\bibitem{Pano3D}
Georgios Albanis, Nikolaos Zioulis, Petros Drakoulis, Vasileios Gkitsas,
  Vladimiros Sterzentsenko, Federico Alvarez, Dimitrios Zarpalas, and Petros
  Daras.
\newblock Pano3d: A holistic benchmark and a solid baseline for 360deg depth
  estimation.
\newblock In {\em Proceedings of the IEEE/CVF Conference on Computer Vision and
  Pattern Recognition}, pages 3727--3737, 2021.

\bibitem{stanford}
Iro Armeni, Sasha Sax, Amir~R Zamir, and Silvio Savarese.
\newblock Joint 2d-3d-semantic data for indoor scene understanding.
\newblock {\em arXiv preprint arXiv:1702.01105}, 2017.

\bibitem{sample_softmax1}
Yoshua Bengio and Jean-S{\'e}bastien Sen{\'e}cal.
\newblock Adaptive importance sampling to accelerate training of a neural
  probabilistic language model.
\newblock {\em IEEE Transactions on Neural Networks}, 19(4):713--722, 2008.

\bibitem{sample_softmax3}
Guy Blanc and Steffen Rendle.
\newblock Adaptive sampled softmax with kernel based sampling.
\newblock In {\em International Conference on Machine Learning}, pages
  590--599. PMLR, 2018.

\bibitem{matterport}
Angel Chang, Angela Dai, Thomas Funkhouser, Maciej Halber, Matthias Niessner,
  Manolis Savva, Shuran Song, Andy Zeng, and Yinda Zhang.
\newblock {Matterport3D}: Learning from {RGB-D} data in indoor environments.
\newblock {\em International Conference on 3D Vision (3DV)}, 2017.

\bibitem{crossvit}
Chun-Fu~Richard Chen, Quanfu Fan, and Rameswar Panda.
\newblock Crossvit: Cross-attention multi-scale vision transformer for image
  classification.
\newblock In {\em Proceedings of the IEEE/CVF international conference on
  computer vision}, pages 357--366, 2021.

\bibitem{ordinal_loss}
Weifeng Chen, Zhao Fu, Dawei Yang, and Jia Deng.
\newblock Single-image depth perception in the wild.
\newblock {\em Advances in neural information processing systems}, 29:730--738,
  2016.

\bibitem{cubepadding}
Hsien-Tzu Cheng, Chun-Hung Chao, Jin-Dong Dong, Hao-Kai Wen, Tyng-Luh Liu, and
  Min Sun.
\newblock Cube padding for weakly-supervised saliency prediction in 360 videos.
\newblock In {\em Proceedings of the IEEE Conference on Computer Vision and
  Pattern Recognition}, pages 1420--1429, 2018.

\bibitem{HA_net}
Sungha Choi, Joanne~T Kim, and Jaegul Choo.
\newblock Cars can't fly up in the sky: Improving urban-scene segmentation via
  height-driven attention networks.
\newblock In {\em Proceedings of the IEEE/CVF conference on computer vision and
  pattern recognition}, pages 9373--9383, 2020.

\bibitem{spherical_cnns}
Taco~S Cohen, Mario Geiger, Jonas K{\"o}hler, and Max Welling.
\newblock Spherical cnns.
\newblock In {\em International Conference on Learning Representations}, 2019.

\bibitem{deformable_conv}
Jifeng Dai, Haozhi Qi, Yuwen Xiong, Yi Li, Guodong Zhang, Han Hu, and Yichen
  Wei.
\newblock Deformable convolutional networks.
\newblock In {\em Proceedings of the IEEE international conference on computer
  vision}, pages 764--773, 2017.

\bibitem{360align}
Benjamin Davidson, Mohsan~S Alvi, and Jo{\~a}o~F Henriques.
\newblock 360 camera alignment via segmentation.
\newblock In {\em Computer Vision--ECCV 2020: 16th European Conference,
  Glasgow, UK, August 23--28, 2020, Proceedings, Part XXVIII 16}, pages
  579--595. Springer, 2020.

\bibitem{Cswin}
Xiaoyi Dong, Jianmin Bao, Dongdong Chen, Weiming Zhang, Nenghai Yu, Lu Yuan,
  Dong Chen, and Baining Guo.
\newblock Cswin transformer: A general vision transformer backbone with
  cross-shaped windows.
\newblock In {\em Proceedings of the IEEE/CVF Conference on Computer Vision and
  Pattern Recognition}, pages 12124--12134, 2022.

\bibitem{vision_transformer}
Alexey Dosovitskiy, Lucas Beyer, Alexander Kolesnikov, Dirk Weissenborn,
  Xiaohua Zhai, Thomas Unterthiner, Mostafa Dehghani, Matthias Minderer, Georg
  Heigold, Sylvain Gelly, et~al.
\newblock An image is worth 16x16 words: Transformers for image recognition at
  scale.
\newblock In {\em International Conference on Learning Representations}, 2021.

\bibitem{silog_loss}
David Eigen, Christian Puhrsch, and Rob Fergus.
\newblock Depth map prediction from a single image using a multi-scale deep
  network.
\newblock {\em Advances in Neural Information Processing Systems}, 27, 2014.

\bibitem{structural_sonar}
Isaac~D Gerg and Vishal Monga.
\newblock Structural prior driven regularized deep learning for sonar image
  classification.
\newblock {\em IEEE Transactions on Geoscience and Remote Sensing}, 60:1--16,
  2021.

\bibitem{TNT}
Kai Han, An Xiao, Enhua Wu, Jianyuan Guo, Chunjing Xu, and Yunhe Wang.
\newblock Transformer in transformer.
\newblock {\em Advances in Neural Information Processing Systems},
  34:15908--15919, 2021.

\bibitem{dilated_attention}
Ali Hassani and Humphrey Shi.
\newblock Dilated neighborhood attention transformer.
\newblock {\em arXiv preprint arXiv:2209.15001}, 2022.

\bibitem{dilated_attention2}
Jiayu Jiao, Yu-Ming Tang, Kun-Yu Lin, Yipeng Gao, Jinhua Ma, Yaowei Wang, and
  Wei-Shi Zheng.
\newblock Dilateformer: Multi-scale dilated transformer for visual recognition.
\newblock {\em IEEE Transactions on Multimedia}, 2023.

\bibitem{geo_depth}
Lei Jin, Yanyu Xu, Jia Zheng, Junfei Zhang, Rui Tang, Shugong Xu, Jingyi Yu,
  and Shenghua Gao.
\newblock Geometric structure based and regularized depth estimation from 360
  indoor imagery.
\newblock In {\em Proceedings of the IEEE/CVF Conference on Computer Vision and
  Pattern Recognition}, pages 889--898, 2020.

\bibitem{depthformer}
Zhenyu Li, Zehui Chen, Xianming Liu, and Junjun Jiang.
\newblock Depthformer: Exploiting long-range correlation and local information
  for accurate monocular depth estimation.
\newblock {\em arXiv preprint arXiv:2203.14211}, 2022.

\bibitem{Swin_v2}
Ze Liu, Han Hu, Yutong Lin, Zhuliang Yao, Zhenda Xie, Yixuan Wei, Jia Ning, Yue
  Cao, Zheng Zhang, Li Dong, et~al.
\newblock Swin transformer v2: Scaling up capacity and resolution.
\newblock In {\em Proceedings of the IEEE/CVF conference on computer vision and
  pattern recognition}, pages 12009--12019, 2022.

\bibitem{SwinT}
Ze Liu, Yutong Lin, Yue Cao, Han Hu, Yixuan Wei, Zheng Zhang, Stephen Lin, and
  Baining Guo.
\newblock Swin transformer: Hierarchical vision transformer using shifted
  windows.
\newblock In {\em Proceedings of the IEEE/CVF International Conference on
  Computer Vision}, pages 10012--10022, 2021.

\bibitem{Lowcost}
Kevin Matzen, Michael~F Cohen, Bryce Evans, Johannes Kopf, and Richard
  Szeliski.
\newblock Low-cost 360 stereo photography and video capture.
\newblock {\em ACM Transactions on Graphics (TOG)}, 36(4):1--12, 2017.

\bibitem{EBS}
Greire Payen~de La~Garanderie, Amir Atapour~Abarghouei, and Toby~P Breckon.
\newblock Eliminating the blind spot: Adapting 3d object detection and
  monocular depth estimation to 360 panoramic imagery.
\newblock In {\em Proceedings of the European Conference on Computer Vision
  (ECCV)}, pages 789--807, 2018.

\bibitem{Slicenet}
Giovanni Pintore, Marco Agus, Eva Almansa, Jens Schneider, and Enrico Gobbetti.
\newblock Slicenet: deep dense depth estimation from a single indoor panorama
  using a slice-based representation.
\newblock In {\em Proceedings of the IEEE/CVF Conference on Computer Vision and
  Pattern Recognition}, pages 11536--11545, 2021.

\bibitem{cosformer}
Zhen Qin, Weixuan Sun, Hui Deng, Dongxu Li, Yunshen Wei, Baohong Lv, Junjie
  Yan, Lingpeng Kong, and Yiran Zhong.
\newblock cosformer: Rethinking softmax in attention.
\newblock In {\em International Conference on Learning Representations}, 2022.

\bibitem{DPT}
Ren{\'e} Ranftl, Alexey Bochkovskiy, and Vladlen Koltun.
\newblock Vision transformers for dense prediction.
\newblock In {\em Proceedings of the IEEE/CVF international conference on
  computer vision}, pages 12179--12188, 2021.

\bibitem{midas}
Ren{\'e} Ranftl, Katrin Lasinger, David Hafner, Konrad Schindler, and Vladlen
  Koltun.
\newblock Towards robust monocular depth estimation: Mixing datasets for
  zero-shot cross-dataset transfer.
\newblock {\em IEEE transactions on pattern analysis and machine intelligence},
  44(3):1623--1637, 2020.

\bibitem{sample_softmax2}
Ankit~Singh Rawat, Jiecao Chen, Felix Xinnan~X Yu, Ananda~Theertha Suresh, and
  Sanjiv Kumar.
\newblock Sampled softmax with random fourier features.
\newblock {\em Advances in Neural Information Processing Systems}, 32, 2019.

\bibitem{Unet}
Olaf Ronneberger, Philipp Fischer, and Thomas Brox.
\newblock U-net: Convolutional networks for biomedical image segmentation.
\newblock In {\em Medical Image Computing and Computer-Assisted
  Intervention--MICCAI 2015: 18th International Conference, Munich, Germany,
  October 5-9, 2015, Proceedings, Part III 18}, pages 234--241. Springer, 2015.

\bibitem{panoformer}
Zhijie Shen, Chunyu Lin, Kang Liao, Lang Nie, Zishuo Zheng, and Yao Zhao.
\newblock Panoformer: Panorama transformer for indoor 360 depth estimation.
\newblock In {\em Computer Vision--ECCV 2022: 17th European Conference, Tel
  Aviv, Israel, October 23--27, 2022, Proceedings, Part I}, pages 195--211.
  Springer, 2022.

\bibitem{spherical_convolution}
Yu-Chuan Su and Kristen Grauman.
\newblock Learning spherical convolution for fast features from 360 imagery.
\newblock {\em Advances in Neural Information Processing Systems}, 30, 2017.

\bibitem{horizonnet}
Cheng Sun, Chi-Wei Hsiao, Min Sun, and Hwann-Tzong Chen.
\newblock Horizonnet: Learning room layout with 1d representation and pano
  stretch data augmentation.
\newblock In {\em Proceedings of the IEEE/CVF Conference on Computer Vision and
  Pattern Recognition}, pages 1047--1056, 2019.

\bibitem{hohonet}
Cheng Sun, Min Sun, and Hwann-Tzong Chen.
\newblock Hohonet: 360 indoor holistic understanding with latent horizontal
  features.
\newblock In {\em Proceedings of the IEEE/CVF Conference on Computer Vision and
  Pattern Recognition}, pages 2573--2582, 2021.

\bibitem{linear_softmax}
Apoorv Vyas, Angelos Katharopoulos, and Fran{\c{c}}ois Fleuret.
\newblock Fast transformers with clustered attention.
\newblock {\em Advances in Neural Information Processing Systems},
  33:21665--21674, 2020.

\bibitem{NMG_loss}
Chaoyang Wang, Simon Lucey, Federico Perazzi, and Oliver Wang.
\newblock Web stereo video supervision for depth prediction from dynamic
  scenes.
\newblock In {\em 2019 International Conference on 3D Vision (3DV)}, pages
  348--357. IEEE, 2019.

\bibitem{360self}
Fu-En Wang, Hou-Ning Hu, Hsien-Tzu Cheng, Juan-Ting Lin, Shang-Ta Yang, Meng-Li
  Shih, Hung-Kuo Chu, and Min Sun.
\newblock Self-supervised learning of depth and camera motion from 360 videos.
\newblock In {\em Asian Conference on Computer Vision}, pages 53--68. Springer,
  2018.

\bibitem{bifuse}
Fu-En Wang, Yu-Hsuan Yeh, Min Sun, Wei-Chen Chiu, and Yi-Hsuan Tsai.
\newblock Bifuse: Monocular 360 depth estimation via bi-projection fusion.
\newblock In {\em Proceedings of the IEEE/CVF Conference on Computer Vision and
  Pattern Recognition}, pages 462--471, 2020.

\bibitem{bifuse_plus}
Fu-En Wang, Yu-Hsuan Yeh, Yi-Hsuan Tsai, Wei-Chen Chiu, and Min Sun.
\newblock Bifuse++: Self-supervised and efficient bi-projection fusion for 360
  depth estimation.
\newblock {\em IEEE Transactions on Pattern Analysis and Machine Intelligence},
  2022.

\bibitem{360sdnet}
Ning-Hsu Wang, Bolivar Solarte, Yi-Hsuan Tsai, Wei-Chen Chiu, and Min Sun.
\newblock 360sdnet: 360 stereo depth estimation with learnable cost volume.
\newblock In {\em 2020 IEEE International Conference on Robotics and Automation
  (ICRA)}, pages 582--588. IEEE, 2020.

\bibitem{pyramidformer}
Wenhai Wang, Enze Xie, Xiang Li, Deng-Ping Fan, Kaitao Song, Ding Liang, Tong
  Lu, Ping Luo, and Ling Shao.
\newblock Pyramid vision transformer: A versatile backbone for dense prediction
  without convolutions.
\newblock In {\em Proceedings of the IEEE/CVF International Conference on
  Computer Vision}, pages 568--578, 2021.

\bibitem{pyramidformer2}
Wenhai Wang, Enze Xie, Xiang Li, Deng-Ping Fan, Kaitao Song, Ding Liang, Tong
  Lu, Ping Luo, and Ling Shao.
\newblock Pvt v2: Improved baselines with pyramid vision transformer.
\newblock {\em Computational Visual Media}, 8(3):415--424, 2022.

\bibitem{non_local}
Xiaolong Wang, Ross Girshick, Abhinav Gupta, and Kaiming He.
\newblock Non-local neural networks.
\newblock In {\em Proceedings of the IEEE conference on computer vision and
  pattern recognition}, pages 7794--7803, 2018.

\bibitem{DAT}
Zhuofan Xia, Xuran Pan, Shiji Song, Li~Erran Li, and Gao Huang.
\newblock Vision transformer with deformable attention.
\newblock In {\em Proceedings of the IEEE/CVF Conference on Computer Vision and
  Pattern Recognition}, pages 4794--4803, 2022.

\bibitem{position_gan}
Rui Xu, Xintao Wang, Kai Chen, Bolei Zhou, and Chen~Change Loy.
\newblock Positional encoding as spatial inductive bias in gans.
\newblock In {\em Proceedings of the IEEE/CVF Conference on Computer Vision and
  Pattern Recognition}, pages 13569--13578, 2021.

\bibitem{trans_depth}
Guanglei Yang, Hao Tang, Mingli Ding, Nicu Sebe, and Elisa Ricci.
\newblock Transformer-based attention networks for continuous pixel-wise
  prediction.
\newblock In {\em Proceedings of the IEEE/CVF International Conference on
  Computer vision}, pages 16269--16279, 2021.

\bibitem{struct_inpainting}
Jie Yang, Zhiquan Qi, and Yong Shi.
\newblock Learning to incorporate structure knowledge for image inpainting.
\newblock In {\em Proceedings of the AAAI Conference on Artificial
  Intelligence}, volume~34, pages 12605--12612, 2020.

\bibitem{dula}
Shang-Ta Yang, Fu-En Wang, Chi-Han Peng, Peter Wonka, Min Sun, and Hung-Kuo
  Chu.
\newblock Dula-net: A dual-projection network for estimating room layouts from
  a single rgb panorama.
\newblock In {\em Proceedings of the IEEE Conference on Computer Vision and
  Pattern Recognition}, pages 3363--3372, 2019.

\bibitem{landmark_face}
Yang Yang and Xiaojie Guo.
\newblock Generative landmark guided face inpainting.
\newblock In {\em Pattern Recognition and Computer Vision: Third Chinese
  Conference, PRCV 2020, Nanjing, China, October 16--18, 2020, Proceedings,
  Part I 3}, pages 14--26. Springer, 2020.

\bibitem{dilated_convolution}
Fisher Yu and Vladlen Koltun.
\newblock Multi-scale context aggregation by dilated convolutions.
\newblock In {\em ICLR}, 2016.

\bibitem{T2T_ViT}
Li Yuan, Yunpeng Chen, Tao Wang, Weihao Yu, Yujun Shi, Zi-Hang Jiang,
  Francis~EH Tay, Jiashi Feng, and Shuicheng Yan.
\newblock Tokens-to-token vit: Training vision transformers from scratch on
  imagenet.
\newblock In {\em Proceedings of the IEEE/CVF international conference on
  computer vision}, pages 558--567, 2021.

\bibitem{joint_360depth}
Ilwi Yun, Hyuk-Jae Lee, and Chae~Eun Rhee.
\newblock Improving 360 monocular depth estimation via non-local dense
  prediction transformer and joint supervised and self-supervised learning.
\newblock In {\em Proceedings of the AAAI Conference on Artificial
  Intelligence}, volume~36, pages 3224--3233, 2022.

\bibitem{jointdepth}
Wei Zeng, Sezer Karaoglu, and Theo Gevers.
\newblock Joint 3d layout and depth prediction from a single indoor panorama
  image.
\newblock In {\em European Conference on Computer Vision}, pages 666--682.
  Springer, 2020.

\bibitem{structure3d}
Jia Zheng, Junfei Zhang, Jing Li, Rui Tang, Shenghua Gao, and Zihan Zhou.
\newblock Structured3d: A large photo-realistic dataset for structured 3d
  modeling.
\newblock In {\em Computer Vision--ECCV 2020: 16th European Conference,
  Glasgow, UK, August 23--28, 2020, Proceedings, Part IX 16}, pages 519--535.
  Springer, 2020.

\bibitem{deformable_conv2}
Xizhou Zhu, Han Hu, Stephen Lin, and Jifeng Dai.
\newblock Deformable convnets v2: More deformable, better results.
\newblock In {\em Proceedings of the IEEE/CVF conference on computer vision and
  pattern recognition}, pages 9308--9316, 2019.

\bibitem{svsyn}
Nikolaos Zioulis, Antonis Karakottas, Dimitrios Zarpalas, Federico Alvarez, and
  Petros Daras.
\newblock Spherical view synthesis for self-supervised 360 depth estimation.
\newblock In {\em 2019 International Conference on 3D Vision (3DV)}, pages
  690--699. IEEE, 2019.

\bibitem{omnidepth}
Nikolaos Zioulis, Antonis Karakottas, Dimitrios Zarpalas, and Petros Daras.
\newblock Omnidepth: Dense depth estimation for indoors spherical panoramas.
\newblock In {\em Proceedings of the European Conference on Computer Vision
  (ECCV)}, pages 448--465, 2018.

\end{thebibliography}
}

\end{document}